\pdfoutput=1
\documentclass[11pt,letterpaper]{article}

\usepackage{iftex}
\usepackage[T1]{fontenc}
\ifPDFTeX
  \usepackage[utf8]{inputenc}
  \input{glyphtounicode}
\fi
\usepackage{lmodern}
\usepackage[margin=1in]{geometry}
\usepackage{amsmath,amssymb}
\usepackage{graphicx}
\usepackage{booktabs,array,longtable,threeparttable,threeparttablex,pdflscape}
\usepackage{siunitx}
\usepackage[font=small,labelfont=bf,skip=4pt]{caption}
\usepackage{enumitem}
\usepackage{needspace}
\usepackage{placeins}
\usepackage{xcolor}
\usepackage{tikz}
\usetikzlibrary{arrows.meta,decorations.pathreplacing,patterns}
\usepackage[round,authoryear]{natbib}
\usepackage{microtype}
\usepackage{xurl}  
\usepackage[hyperfootnotes=false]{hyperref}
\hypersetup{
  pdftitle={Thinking Fast, Trading Late: System One Decision Models vs the Market},
  pdfauthor={Amir Dehkordi, Alireza Kargarzadeh, Nariman Khaledian, Navid Parvini, Sid Ghatak, Arman Khaledian},
  pdfsubject={Large language models and AI decision models on financial news: a pre-registered,
    look-ahead-free test of stock-price reactions},
  pdfkeywords={large language models (LLMs); AI decision models; stock return prediction; financial news;
    look-ahead bias; memorization; data contamination; backtest overfitting; deflated Sharpe ratio;
    pre-registration; overnight returns; market efficiency; overreaction and underreaction; calibration},
  colorlinks=true,
  linkcolor={blue!50!black},
  citecolor={blue!50!black},
  urlcolor={blue!50!black}
}
\newcommand{\figdir}{figures}

\makeatletter
\DeclareRobustCommand{\res}[1]{\@ifundefined{res@#1}{\PackageError{numbers}{Unknown result `#1'}{Add it to paper/numbers.yaml and run make paper-text.}}{\csname res@#1\endcsname}}
\@namedef{res@ntrials-strat}{545}
\@namedef{res@ntrials-total}{1{,}620}
\@namedef{res@test-d0-events}{13{,}748}
\@namedef{res@test-d0-sessions}{140}
\@namedef{res@test-d2-events}{17{,}188}
\@namedef{res@valb-sessions}{59}
\@namedef{res@f5-recall}{43.5\%}
\@namedef{res@ablation-failed}{210}
\@namedef{res@lf-ic-vala}{0.055}
\@namedef{res@lf-ic-valb}{0.065}
\@namedef{res@lf-ic-valb-t}{3.64}
\@namedef{res@lf-ic-valb-t1}{3.6}
\@namedef{res@lf-ic-valb-lo}{0.030}
\@namedef{res@lf-ic-test}{0.020}
\@namedef{res@lf-ic-test-t}{1.67}
\@namedef{res@lf-ic-test-t1}{1.7}
\@namedef{res@d1-ic-valb}{0.097}
\@namedef{res@d1-ic-test}{0.039}
\@namedef{res@llm-ic-valb}{0.052}
\@namedef{res@llm-ic-test}{0.001}
\@namedef{res@tags-ic-test}{0.055}
\@namedef{res@lf-h1-holm}{0.853}
\@namedef{res@lf-h3-holm}{0.269}
\@namedef{res@lf-h3}{0.0038}
\@namedef{res@lf-h3-p}{0.054}
\@namedef{res@lf-h4}{0.0012}
\@namedef{res@lf-h4-p}{0.216}
\@namedef{res@lf-auc}{0.60}
\@namedef{res@d1-ic-on-test}{0.074}
\@namedef{res@span-ic-on-test-t}{4.49}
\@namedef{res@drex-ic-on-test}{0.077}
\@namedef{res@drex-ic-on-test-t}{5.34}
\@namedef{res@d1-ic-id-test}{\ensuremath{-}0.001}
\@namedef{res@drex-ic-id-test}{0.002}
\@namedef{res@lz-ic-id-test}{\ensuremath{-}0.017}
\@namedef{res@d1-h1}{0.075}
\@namedef{res@d1-h1-lo}{0.040}
\@namedef{res@d1-h1-hi}{0.109}
\@namedef{res@d1-h1-p}{0.0001}
\@namedef{res@d1-qspread-on}{0.81}
\@namedef{res@d1-qspread-on-t}{5.6}
\@namedef{res@lz-h1}{0.054}
\@namedef{res@lz-h1-p}{0.001}
\@namedef{res@lf-ic-on-test}{0.015}
\@namedef{res@lf-ic-id-test}{0.010}
\@namedef{res@lf-ic-id-test-t}{0.8}
\@namedef{res@lf-ic-id-valb}{0.054}
\@namedef{res@lf-ic-id-valb-t}{2.89}
\@namedef{res@d1-ic-on-0815}{0.115}
\@namedef{res@d1-ic-on-0815-t}{5.0}
\@namedef{res@d1-ic-on-0408}{0.061}
\@namedef{res@d1-ic-on-pre04}{0.024}
\@namedef{res@d1-h1-pre04}{0.022}
\@namedef{res@d1-h1-0815}{0.122}
\@namedef{res@unmask-n}{5{,}923}
\@namedef{res@unmask-ic-masked}{0.026}
\@namedef{res@unmask-ic-unmasked}{0.029}
\@namedef{res@lf-h2-b-d2}{\ensuremath{-}0.044}
\@namedef{res@lf-h2-bp-d2}{0.62}
\@namedef{res@lf-h2-c-d2}{\ensuremath{-}0.015}
\@namedef{res@lf-h2-cp-d2}{0.11}
\@namedef{res@lf-h2-n-d2}{17{,}188}
\@namedef{res@d1-h2-b-d2}{\ensuremath{-}0.076}
\@namedef{res@d1-h2-bp-d2}{0.99}
\@namedef{res@het-q-min}{0.74}
\@namedef{res@lf-kappa-ct}{0.36}
\@namedef{res@lf-kappa-t20}{0.26}
\@namedef{res@lf-kappa-incr-p}{0.47}
\@namedef{res@d1-kappa-ct}{0.59}
\@namedef{res@d1-kappa-t20}{0.36}
\@namedef{res@d1-kappa-incr-p}{0.83}
\@namedef{res@d1-dec10-pre5}{\ensuremath{-}1.19}
\@namedef{res@d1-dec1-pre5}{1.54}
\@namedef{res@lf-lambda-t1}{\ensuremath{-}0.007}
\@namedef{res@lf-lambda-t5}{0.026}
\@namedef{res@lf-lambda-t5-p}{0.16}
\@namedef{res@lf-lambda-t20}{0.050}
\@namedef{res@lf-tracking-t20}{1.001}
\@namedef{res@lf-tracking-t20-p}{0.87}
\@namedef{res@rg-under-n}{5{,}021}
\@namedef{res@rg-under-t20}{\ensuremath{-}1.04}
\@namedef{res@rg-under-v}{0.38}
\@namedef{res@rg-over-t0}{3.16}
\@namedef{res@rg-over-t20}{2.82}
\@namedef{res@drift-g3-medium}{\ensuremath{-}0.18}
\@namedef{res@drift-g3-weak}{\ensuremath{-}0.09}
\@namedef{res@lf-kappa-ct-top100}{0.75}
\@namedef{res@lf-kappa-ct-mid}{0.23}
\@namedef{res@lf-kappa-ct-rest}{0.25}
\@namedef{res@d1-kappa-t20-numbers}{0.78}
\@namedef{res@d1-kappa-t20-numbers-p}{0.002}
\@namedef{res@d1-kappa-t20-narrative}{\ensuremath{-}0.04}
\@namedef{res@roll-r2-pct}{25.3\%}
\@namedef{res@lf-ladder-post-pct}{0.25\%}
\@namedef{res@d1-ladder-post-pct}{0.82\%}
\@namedef{res@d1-ladder-expl-pct}{15.8\%}
\@namedef{res@d1-ladder-expl-event-pct}{37.5\%}
\@namedef{res@lf-h5-q}{0.058}
\@namedef{res@theta-d2}{\ensuremath{-}0.062}
\@namedef{res@theta-d2-p}{0.95}
\@namedef{res@lf-ece}{0.057}
\@namedef{res@lf-llskill}{0.001}
\@namedef{res@span-ece}{0.361}
\@namedef{res@drex-ece}{0.209}
\@namedef{res@d1-ece-before}{0.260}
\@namedef{res@d1-ece-after}{0.155}
\@namedef{res@mem-n}{400}
\@namedef{res@mem-baseline}{28\%}
\@namedef{res@mem-baseline-f2}{0.28}
\@namedef{res@mem-baseline-na06}{36\%}
\@namedef{res@mem-min}{62\%}
\@namedef{res@mem-max}{74\%}
\@namedef{res@mem-span}{66\%}
\@namedef{res@mem-drex}{68\%}
\@namedef{res@mem-laya}{32\%}
\@namedef{res@rel-laya}{2026-09-18}
\@namedef{res@rel-drex}{2026-10-12}
\@namedef{res@rel-span}{2026-10-16}
\@namedef{res@sconv-ndays}{139}
\@namedef{res@sconv-shared}{92\%}
\@namedef{res@sconv-corr}{0.96}
\@namedef{res@sd0-gated-gross}{1.88}
\@namedef{res@sd0-gated-be}{8.3}
\@namedef{res@sd0-turnover}{4.4}
\@namedef{res@sd2j-pct}{3}
\@namedef{res@dsr-max}{0.024}
\@namedef{res@dsr-sr0}{0.26}
\@namedef{res@earn-n-test}{1{,}128}
\@namedef{res@lf-earn-vala}{0.125}
\@namedef{res@lf-earn-valb}{0.106}
\@namedef{res@lf-earn-test}{0.134}
\@namedef{res@lf-rest-test}{0.013}
\@namedef{res@d1-earn-test}{0.223}
\@namedef{res@d1-earn-test-t}{8.1}
\@namedef{res@d1-earn-on-test}{0.238}
\@namedef{res@d1-earn-id-test}{0.034}
\@namedef{res@sel-d1-on-10-vala}{1.52}
\@namedef{res@sel-d1-on-10-valb}{1.16}
\@namedef{res@sel-d1-on-10-test}{0.93}
\@namedef{res@sel-d1-on-all-vala}{0.22}
\@namedef{res@sel-d1-on-all-valb}{0.26}
\@namedef{res@sel-d1-on-all-test}{0.21}
\@namedef{res@watch-lf-rev-test}{\ensuremath{-}0.036}
\@namedef{res@watch-lf-rev-test-t}{\ensuremath{-}2.3}
\@namedef{res@cond-pairs}{75}
\@namedef{res@cond-selected}{8}
\@namedef{res@cond-reversed}{5}
\@namedef{res@cond-positive}{3}
\@namedef{res@cb-valb-gross-max}{6.8}
\makeatother

\newcommand{\Nhat}{\hat N}
\newcommand{\Ntilde}{\tilde N}
\newcommand{\IC}{\mathrm{IC}}
\newcommand{\AR}{\mathrm{AR}}
\newcommand{\CARlvl}{\mathrm{CAR}^{\mathrm{lvl}}}
\newcommand{\Glvl}{G^{\mathrm{lvl}}}

\newcommand{\authorinfo}[3]{\\{\small #1}\\{\small #2}\\{\small\href{mailto:#3}{#3}}}
\newcommand{\authorrow}{\end{tabular}\par\vskip 1em\begin{tabular}[t]{c}}

\title{Thinking Fast, Trading Late:\\ System One Decision Models vs the Market%
  \thanks{Code availability: \url{https://github.com/AmirDehkordi/NewsAction}. The data are private
  and are not redistributed (Appendix~\ref{app:replication}).}}
\author{%
  Amir Dehkordi\thanks{Corresponding author.}%
  \authorinfo{Independent Researcher}{London, United Kingdom}{amir.dehkordi98@gmail.com}
  \and Alireza Kargarzadeh%
  \authorinfo{TailState Intelligence Ltd}{London, United Kingdom}{alireza.kargarzadeh@tailstate.ai}
  \and Nariman Khaledian%
  \authorinfo{Independent Researcher}{Antibes, France}{nariman.khaledian@gmail.com}
  \authorrow Navid Parvini%
  \authorinfo{Zanista AI Ltd}{London, United Kingdom}{navid.parvini@zanista.ai}
  \and Sid Ghatak%
  \authorinfo{Increase Alpha, LLC}{Miami, United States}{s.ghatak@increasealpha.com}
  \and Arman Khaledian%
  \authorinfo{Zanista AI Ltd}{London, United Kingdom}{arman.khaledian@zanista.ai}}
\date{November 2026}

\begin{document}

\maketitle
\thispagestyle{empty}

\begin{abstract}
\noindent
Large language models (LLMs) read company news at scale, and backtests suggest that their readings predict stock
returns. We ask how much of that predictability survives a test that rules out look-ahead, memorization and
selection. We propose OLIVIA, a protocol for such tests: point-in-time inputs with company identities masked,
selection on validation data only, an analysis plan frozen by cryptographic hash, and a test period unblinded once
through a logged lock. We apply it to AI decision models, non-generative language models that return
probabilities for typed questions, reading news about S\&P 500 stocks: an open-weights model we fine-tuned on
2024--2025 data, three commercial models and standard baselines. The fine-tuned model's signal replicated on two
validation quarters (daily information coefficients \res{lf-ic-vala} and \res{lf-ic-valb}) but fell to
\res{lf-ic-test} ($t = \res{lf-ic-test-t1}$) on the blinded January--July 2026 test. The commercial models'
readings line up with the overnight move to the open (information coefficient \res{d1-ic-on-test} to
\res{drex-ic-on-test}) but not with the move after it, so the news is priced by the time a trader can act on it at
the open. No model can tell over- from under-reaction, and no pre-registered strategy survives costs and a deflated
Sharpe ratio. Commercial models identify masked companies in \res{mem-min} to \res{mem-max} of cases against a
\res{mem-baseline} baseline: masking hides names, not companies. The market reads the news first.

\end{abstract}

\noindent\textbf{Keywords:} large language models (LLMs); financial news; stock return prediction; look-ahead
bias; memorization; pre-registration; overnight returns; market efficiency.

\medskip
\noindent\textbf{JEL codes:} G14, G12, C45, C52, C12.

\clearpage

\section{Introduction}\label{sec:intro}

Much company news is published before the stock market opens. Two questions follow. Can a model that reads that
news anticipate how the price will react? And once part of the reaction is visible, can it tell whether the
market moved too far or not far enough? Both are old questions in empirical finance. Prices react to firm news
within minutes \citep{patell1984,busse2002}, media tone predicts short-horizon returns \citep{tetlock2007}, and
moves that come with news tend to drift while moves without news tend to reverse \citep{chan2003}. Large
language models (LLMs) now read news at scale, and several studies find that their readings predict returns
\citep{lopezlira2026,chen2022}. We ask whether such predictability survives a test built to rule out the usual
ways a backtest misleads: look-ahead in the data, memorization in the model and selection over many variants.

We study \emph{AI decision models}. A decision model is a non-generative model that takes a \emph{state} (text
plus structured context) and a set of typed questions (a choice among options, a score on an ordered scale, or
a yes-or-no question) and returns probabilities over the answers instead of generating text
\citep{zhu2026}. Decision models are close relatives of LLM-based classifiers and of probabilistic
forecasting with language models \citep{kadavath2022,halawi2024,schoenegger2024}, and their vendors describe the
probabilities as calibrated. They suit finance because they return the quantities a forecaster needs, such as
a distribution over the size of the coming move, at a cost of a few cents per thousand decisions, and because
open-weights versions can be fine-tuned.

Language models also bring a specific threat. A model trained on text published after the period on which it
is tested may know how events turned out, and even a model with an earlier training cutoff can recognize a
company from the content of its news \citep{glasserman2024,sarkar2024,lopezlira2025}. A backtest of a
language-model signal on historical news therefore mixes the model's ability to read news with its memory of
what followed. A second threat is older: when many signals, filters and horizons are tried, the best backtest
overstates what a strategy will earn \citep{harvey2016,bailey2014}. Pre-registration, fixing the analysis before
the data are seen, is the standard remedy in other sciences and is still rare in empirical finance
\citep{harvey2017,olken2015}.

We combine these elements in OLIVIA, a protocol for testing language-model signals on financial news without
look-ahead (Table~\ref{tab:olivia}), and apply it to the point-in-time S\&P 500 from July 2024 to July 2026
(Figure~\ref{fig:design}). For each company and trading session we build the news state as of 09:15~ET and mask
company names, tickers, people, dates and every mention of a price move. Before the open, each model gives a
\emph{pre-open reading}: how the stock will perform once the news is absorbed. At 10:00 and after the close it
gives a \emph{post-reaction reading}: was the reaction so far too small or too large for the news? The primary
instrument, Laya-Fin, is an open-weights decision model that we fine-tuned on 2024--2025 news and returns; we
compare it with three commercial decision models, the untuned Laya model and standard baselines. The hypotheses
and selection rules were written down before any data were analyzed, and every later change is dated in an
append-only log; selection used two validation quarters of 2025, and the analysis plan was frozen by hash before
the 2026 test outcomes were unblinded, once.

\needspace{8\baselineskip}
\paragraph{Key findings.}
\begin{enumerate}[label=(\arabic*), leftmargin=*, itemsep=3pt]
\item \textbf{Pre-open news is priced at the open.} The commercial models' readings line up with the overnight
  move from the previous close to the open (daily information coefficient, IC, \res{d1-ic-on-test} to
  \res{drex-ic-on-test}) but not with the move after the open (about zero). No model predicts the move after the
  open, which is the part a trader who reads the news before the open can still earn. The untuned Laya model,
  which barely recognizes masked companies, shows the same split.
\item \textbf{A validated signal did not survive the blinded test.} Laya-Fin's IC with the day's move was
  \res{lf-ic-valb} ($t = \res{lf-ic-valb-t1}$) in the final validation quarter and \res{lf-ic-test}
  ($t = \res{lf-ic-test-t1}$) on the test. Every model whose validation IC was clearly positive fell, none of the five
  primary hypotheses is supported, and no pre-registered strategy survives costs and a deflated Sharpe ratio. In
  exploratory analyses, the signal that remained sat on earnings days, and conditions selected on two validation
  quarters, with in-sample Sharpe ratios up to \res{cb-valb-gross-max}, failed on the test.
\item \textbf{The models cannot judge the reaction.} In the pre-registered over/under regression the news
  coefficient is not positive (Laya-Fin $b = \res{lf-h2-b-d2}$, $p = \res{lf-h2-bp-d2}$), and the moves the
  models call underreactions never catch up. Out of sample, same-day news explains \res{lf-ladder-post-pct}
  (Laya-Fin) to \res{d1-ladder-post-pct} (d1-ultra) of a stock's own daily variance, and the commercial models'
  stated probabilities overstate their accuracy.
\item \textbf{Masking is not anonymity.} Given masked news, the commercial models pick the company out of four
  same-sector candidates in \res{mem-min} to \res{mem-max} of cases, against a matched baseline of
  \res{mem-baseline}; the open-weights Laya model does so in \res{mem-laya}.
\end{enumerate}

\paragraph{Contributions.} First, we give OLIVIA, a protocol of six safeguards (one-time unblinding,
look-ahead-free states, identity masking, validation-only selection, an immutable plan and an audit trail), which
Table~\ref{tab:olivia} lists as a checklist that other studies can adopt. Second, we give a test of whether a
language-model news signal can be traded: splitting each day's move at the open shows that the predictable part of
pre-open news is already in the opening price. Third, we add two measurement tools for language models in finance:
a matched-distractor test of whether a model recognizes masked companies, and a firm-level version of the
explanation ladder of \citet{kelly2026}, which measures how much of a stock's own daily move its news explains.

Section~\ref{sec:related} reviews related work and Section~\ref{sec:design} describes the data and the design.
Sections~\ref{sec:validation} to~\ref{sec:strategies} report the pre-registered results, nulls included.
Section~\ref{sec:exploratory} reports exploratory analyses, labeled post hoc, and Section~\ref{sec:limits} the
limitations and a pre-registered forward test on news published after the models' release.

\section{Related work}\label{sec:related}

\paragraph{Language models and stock returns.} Text has long been used to predict returns, from dictionary
measures of the tone of news \citep{tetlock2007,tetlock2008} to sentiment scores learned from returns
\citep{ke2019}. Large language models extend this line. \citet{lopezlira2026} find that ChatGPT's reading of
headlines predicts the returns that follow. For news released overnight, it predicts the move from the previous
close to the open, which they note only an insider or a trader with early access could exploit, and a smaller
drift from the open to the close, which they trade and which turns unprofitable at 20 basis points per round trip;
our hypothesis H1b tests that open-to-close leg. \citet{chen2022} find that LLM representations of news predict the
cross-section of returns. \citet{kelly2026} use generative AI to explain realized market returns with the
reasons given in daily market summaries, revisiting \citet{roll1988}, who found that market, industry and
public news explain only a small part of individual stock returns; we carry their explanation ladder to the
firm level. Decision models are a newer interface
\citep{zhu2026}: the model returns probabilities for typed questions instead of text, so no free
text needs parsing and every output is a forecast that can be scored.

\paragraph{Look-ahead bias and memorization.} \citet{glasserman2024} test for look-ahead bias in GPT's
sentiment readings of news by anonymizing the companies; \citet{sarkar2024} analyze look-ahead bias in
pretrained language models used for prediction; \citet{lopezlira2025} show that LLMs recall economic and
financial data from their training period, so forecasting tests inside that period are not out of sample;
\citet{he2026} train language models only on text available at each date; and \citet{ludwig2026} argue that a
language model can be used for prediction only if its training data exclude the evaluation sample, which
undisclosed training data rarely allow a researcher to verify. In machine learning, memorization
and benchmark contamination are documented and measured directly \citep{carlini2023,golchin2024,sainz2023}. Every
decision model we use was released after our sample ended, so none can be assumed ignorant of it. We therefore
mask the states, measure how often a model still identifies the company, and pre-register a forward test on
news published after the models' release.

\paragraph{News, attention and price reactions.} Prices incorporate firm news quickly
\citep{patell1984,busse2002} but not always completely. Earnings news drifts \citep{bernard1989}; price moves
that come with news continue while moves without identifiable news reverse \citep{chan2003,savor2012};
high-frequency evidence points to underreaction to news \citep{jiang2021}; and identifying the relevant news
sharpens the link between information and volatility \citep{boudoukh2019}. Stale or recombined news is
mispriced \citep{tetlock2011,fedyk2023}, and distracted or inattentive investors underreact
\citep{hirshleifer2009,dellavigna2009}. Behavioral models explain both underreaction and overreaction
\citep{barberis1998,daniel1998}, building on the overreaction evidence of \citet{debondt1985}, while
\citet{fama1998} argues that the two kinds of anomaly are about equally frequent and fragile to the choice of
method. The split of a daily return into an overnight and an intraday part \citep{lou2019} is central here: we
ask on which side of the open the news is priced. Much of the price discovery for news released outside regular
hours happens in after-hours and pre-market trading \citep{barclay2003}, and the overnight returns of stocks that
attract attention partly reverse during the trading day \citep{berkman2012}.

\paragraph{Selective prediction and calibration.} A model that can abstain trades coverage for accuracy
\citep{geifman2017}, and whether its confidence is a useful guide depends on calibration, which modern neural
networks often lack \citep{pakdamannaeini2015,guo2017}. Language models are partly calibrated about what they know
\citep{kadavath2022}, and LLM forecasts can approach the accuracy of human forecasters
\citep{halawi2024,schoenegger2024}. We measure calibration with the pre-registered expected calibration error
and use selective prediction only in the exploratory section.

\paragraph{Pre-registration and replication in finance.} Many published return predictors do not survive
multiple-testing adjustments or replication \citep{harvey2016,hou2020}, although \citet{jensen2023} find that
most factors replicate once the evidence is aggregated properly, and the returns to published predictors fall
out of sample and after publication \citep{mclean2016}. \citet{harvey2017} calls for pre-registration
and other safeguards in financial economics, and \citet{olken2015} discusses the benefits and limits of
pre-analysis plans in economics. For trading strategies, the deflated Sharpe ratio adjusts a backtest for the
number of trials behind it \citep{bailey2014}. We pre-register hypotheses, families and corrections, log every
trial, and report every pre-registered result.

\paragraph{Our earlier work.} \citet{kargarzadeh2026} ask when news is priced in, using millions of news
articles on about three thousand US stocks, each tagged by an LLM-distilled classifier with an event type and
several attributes. They find that the reaction is concentrated before and on the day of publication, that
markets underreact to numbers and overreact to stories, and that news carries width as well as direction. This
paper extends that work in three ways. It replaces the event tags with decision models that read the masked
news and forecast the move directly; it splits the reaction at the open and at fixed decision points, so that
it can ask whether a reader of the news can trade on it; and it moves from a descriptive study to a
pre-registered, blinded test with a single unblinding.

\section{Data and design}\label{sec:design}

\begin{figure}[tbp]
\centering
\definecolor{figink}{HTML}{0B0B0B}
\definecolor{figinktwo}{HTML}{52514E}
\definecolor{figmuted}{HTML}{898781}
\definecolor{figband}{HTML}{D6D5CE}
\definecolor{figgrid}{HTML}{E1E0D9}
\begin{tikzpicture}[x=1cm, y=1cm, font=\scriptsize, line cap=butt,
    axis/.style={draw=figinktwo, line width=0.5pt},
    tick/.style={draw=figinktwo, line width=0.5pt},
    lead/.style={draw=figmuted, line width=0.4pt},
    lab/.style={text=figink, inner sep=1pt},
    sub/.style={text=figinktwo, inner sep=1pt},
    brk/.style={draw=figmuted, line width=0.5pt},
    brace/.style={decorate, decoration={brace, mirror, amplitude=3pt}, draw=figinktwo, line width=0.5pt}]
%
\node[anchor=west, font=\footnotesize, inner sep=0] at (0,2.75) {\textbf{(a)} The calendar};
\draw[lead] (11.40,-0.12) -- (11.40,0.95);   
\node[lab, anchor=east] at (11.35,0.95) {Laya released (18 Sep)};
\draw[figinktwo, line width=2pt] (13.20,-0.12) -- (13.50,-0.12);   
\draw[lead] (13.20,-0.12) -- (13.20,1.40);
\node[lab, anchor=east] at (13.15,1.40) {commercial models released (12--16 Oct)};
\draw[lead] (14.10,0.5) -- (14.10,1.85);      
\node[lab, anchor=east] at (14.05,1.85) {pre-registration written (24 Oct)};
\draw[lead] (14.40,0.5) -- (14.40,2.30);      
\node[lab, anchor=east] at (14.35,2.30) {plan frozen, then test unblinded once (28 Oct)};
\fill[figband] (0.00,0) rectangle (4.78,0.5);
\fill[figband] (4.82,0) rectangle (5.98,0.5);
\fill[figband] (6.02,0) rectangle (7.18,0.5);
\fill[white] (7.22,0) rectangle (10.0,0.5);
\fill[pattern=north east lines, pattern color=figband] (7.22,0) rectangle (10.0,0.5);
\draw[figinktwo, line width=0.5pt] (7.22,0) rectangle (10.0,0.5);
\node[lab] at (2.39,0.25) {training};
\node[lab] at (5.40,0.25) {val-A};
\node[lab] at (6.60,0.25) {val-B};
\node[lab, fill=white, inner sep=1.5pt] at (8.61,0.25) {test, sealed};
\fill[white] (13.95,0) rectangle (16.15,0.5);
\draw[figinktwo, line width=0.5pt, dash pattern=on 2pt off 1.5pt] (16.15,0.5) -- (13.95,0.5) -- (13.95,0) -- (16.15,0);
\draw[figinktwo, line width=0.5pt, -{Stealth[length=3.5pt]}] (16.0,0.25) -- (16.4,0.25);
\node[lab] at (15.05,0.25) {forward test};
\draw[axis] (0,-0.12) -- (10.0,-0.12);
\draw[axis] (10.8,-0.12) -- (16.15,-0.12);
\draw[brk] (10.27,-0.24) -- (10.37,0.0);
\draw[brk] (10.43,-0.24) -- (10.53,0.0);
\foreach \x in {0,2.4,4.8,7.2,9.6,12.375,14.70} \draw[tick] (\x,-0.12) -- (\x,-0.21);
\node[sub, anchor=north west] at (0,-0.23) {Jul 2024};
\node[sub, anchor=north] at (2.4,-0.23) {Jan 2025};
\node[sub, anchor=north] at (4.8,-0.23) {Jul 2025};
\node[sub, anchor=north] at (7.2,-0.23) {Jan 2026};
\node[sub, anchor=north] at (9.6,-0.23) {Jul 2026};
\node[sub, anchor=north] at (12.375,-0.23) {1 Oct 2026};
\node[sub, anchor=north] at (14.70,-0.23) {1 Nov 2026};
%
\node[anchor=west, font=\footnotesize, inner sep=0] at (0,-1.2) {\textbf{(b)} One trading session (not to scale)};
\begin{scope}[yshift=-3.05cm]
\fill[figgrid] (0,0.92) rectangle (5.0,1.27);
\node[lab] at (2.5,1.095) {news read at D0};
\draw[lead] (5.0,0.92) -- (5.0,0.42);
\fill[figband] (0,-0.07) rectangle (1.2,0.07);
\fill[figband] (3.0,-0.07) rectangle (6.0,0.07);
\node[sub, anchor=south west] at (0,0.1) {after-hours};
\node[sub, anchor=south west] at (3.0,0.1) {pre-market};
\draw[axis] (0,0) -- (10.4,0);
\draw[axis] (10.7,0) -- (14.3,0);
\draw[axis] (14.6,0) -- (16.0,0);
\draw[brk] (10.47,-0.12) -- (10.57,0.12);
\draw[brk] (14.37,-0.12) -- (14.47,0.12);
\foreach \x in {0,3.0,5.0,6.0,7.2,10.0,11.0,13.3,15.6} \draw[tick] (\x,0) -- (\x,-0.1);
\node[sub, anchor=north west, align=left] at (0,-0.14) {$C_{t-1}$\\16:00};
\node[sub, anchor=north] at (3.0,-0.14) {04:00};
\node[sub, anchor=north] at (5.0,-0.14) {09:15};
\node[sub, anchor=north, align=center] at (6.0,-0.14) {$O_t$\\09:30};
\node[sub, anchor=north] at (7.2,-0.14) {10:00};
\node[sub, anchor=north, align=center] at (10.0,-0.14) {$C_t$\\16:00};
\node[sub, anchor=north] at (11.0,-0.14) {$O_{t+1}$};
\node[sub, anchor=north] at (13.3,-0.14) {$C_{t+5}$};
\node[sub, anchor=north] at (15.6,-0.14) {$C_{t+20}$};
\foreach \x/\d in {5.0/D0, 7.2/D1, 10.0/D2} {
  \fill[figink] (\x,0) circle (0.07);
  \node[lab, anchor=south, font=\scriptsize\bfseries] at (\x,0.12) {\d};
}
\draw[brace] (0,-0.86) -- (6.0,-0.86) node[midway, below=4pt, lab] {overnight move $z_{on}$};
\draw[brace] (6.0,-0.86) -- (10.0,-0.86) node[midway, below=4pt, lab] {move after the open $z_{id}$};
\draw[brace] (11.0,-0.86) -- (13.3,-0.86) node[midway, below=4pt, lab] {next week $y_2$};
\end{scope}
\end{tikzpicture}
\caption{\textbf{The design: every choice was made on 2025 validation data, the 2026 test was sealed until the
plan was frozen, and every model was released after the sample ended.} (a) The calendar; the time axis changes
scale at the break, from months to days. The test outcomes were released only after the freeze. The forward
test, pre-registered on 2026-10-29 and not yet run, uses news published from 2026-10-22 on. (b) One trading
session. At D0 (09:15~ET) a
model reads the news published since the previous close; at D1 (10:00) and D2 (after the close) it also sees the
reaction so far. The overnight move $z_{on}$ runs from the previous close to the open, so it includes after-hours
and pre-market trading; the move after the open $z_{id}$ runs from the open to the close; and $y_2$ runs from the
next open to the close of $t+5$. The convergence ladder of Section~\ref{sec:convergence} runs to
$C_{t+20}$.}\label{fig:design}
\end{figure}
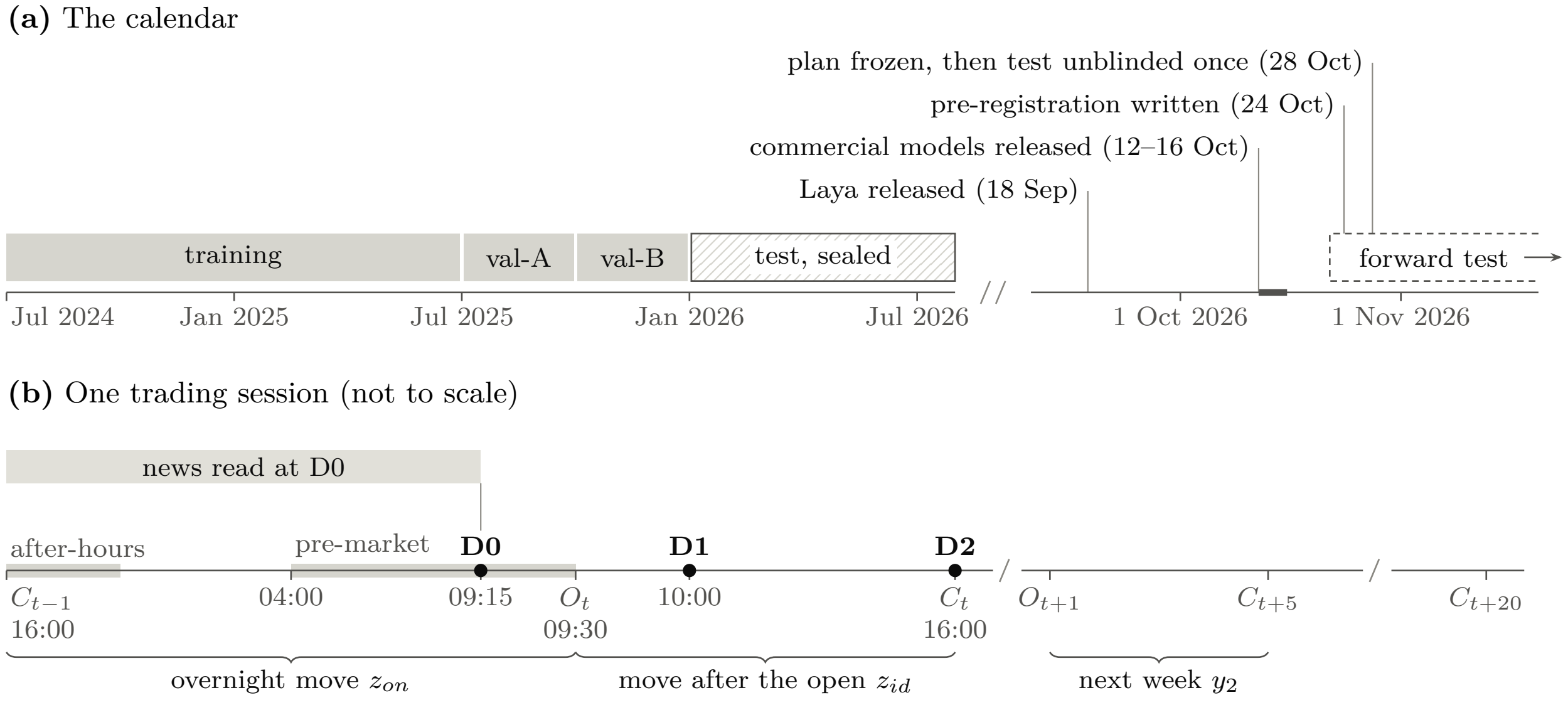

The design implements the OLIVIA protocol (Table~\ref{tab:olivia}); Figure~\ref{fig:design} sets out its calendar
and one trading session. The pre-registration, its amendments and its deviations log are part of the replication
package. Appendix~\ref{app:prereg} summarizes the material deviations, and Appendix Table~\ref{tab:notation}
collects the notation.

\begin{table}[tbp]
\centering
\footnotesize
\setlength{\tabcolsep}{4pt}
\caption{The OLIVIA protocol: six safeguards for testing language-model signals on financial news, what each
guards against, how this study implements it and what a replication needs. The initials of the safeguards give
the name.}\label{tab:olivia}
\begin{tabular}{@{}>{\raggedright\arraybackslash}p{1.0in}>{\raggedright\arraybackslash}p{1.4in}%
>{\raggedright\arraybackslash}p{2.6in}>{\raggedright\arraybackslash}p{1.1in}@{}}
\toprule
Safeguard & Guards against & In this study & A replication needs \\
\midrule
\textbf{O}ne-time unblinding & Repeated looks at the test period & The 2026 prices were sealed behind a lock that
checks the frozen hash; it released the test outcomes only after the freeze, to the frozen analyses, and logs
every read & Sealed outcomes and a log of
every access \\
\addlinespace
\textbf{L}ook-ahead-free states & Inputs that were not public at the decision & Every input carries the time it
became known; the news ends at 09:15~ET, a rounded time stamp counts from the end of its hour or day, and rolling
statistics end the day before & An availability time for every input \\
\addlinespace
\textbf{I}dentity masking & A model recalling what happened to a company it recognizes & Names, tickers, people,
dates and price moves are removed; an independent scanner checks the result; a matched-distractor test measures
recognition (Section~\ref{sec:memorization}) & A masker, a leak check and a recognition test \\
\addlinespace
\textbf{V}alidation-only selection & Choosing models, filters or thresholds on the test & Search and calibration
on val-A; confirmation and the final selection on val-B & Two validation periods before the test \\
\addlinespace
\textbf{I}mmutable plan & Changing hypotheses or tests after seeing results & The pre-registration and the analysis
code were frozen by a SHA-256 hash and a version-control tag before any test-period call; later changes go to a
dated, append-only deviations log & A hash, a tag and a deviations log \\
\addlinespace
\textbf{A}udit trail & Unreported variants & Append-only ledgers of every evaluated variant (their count sets the
bar of the deflated Sharpe ratio), of API spend and of every access to test-period data & A trial ledger kept from
the first analysis \\
\bottomrule
\end{tabular}
\end{table}

\subsection{Universe, sample periods and returns}\label{sec:universe}

The universe is the point-in-time membership of the S\&P 500: a stock is in the sample on a session only if it
was an index member on that session, and companies with several share classes count once. Each month the
members are split into three size tiers by market capitalization at the previous month-end: the 100 largest,
ranks 101--250 and ranks 251--500. The sample is split by time into a training period (2024-07-01 to
2025-06-30), two validation quarters and a test period. The first validation quarter, val-A (2025-07-01 to
2025-09-30), is used for search and for fitting calibration maps, temperatures and thresholds; the second, val-B
(2025-10-01 to 2025-12-31), for confirmation and the final selection. The test period, 2026-01-02 to
2026-07-31, was blinded until the analysis plan was frozen and was then unblinded once. A target enters an
evaluation only if its whole horizon ends inside its period, and the first five sessions of each evaluation
period are left out.

Daily prices are adjusted for splits, dividends and spin-offs, and the opening price is put on the same basis.
For stock $i$ on session $t$, the close-to-close return splits at the open into an overnight return, from the
previous close to the open, and an intraday return, from the open to the close:
\begin{equation}\label{eq:split}
1 + r^{cc}_{i,t} = \bigl(1 + r^{on}_{i,t}\bigr)\bigl(1 + r^{id}_{i,t}\bigr),
\qquad
z^{k}_{i,t} = \frac{\AR^{k}_{i,t}}{\sigma^{k}_{i,t-1}}, \quad k \in \{cc, on, id\}.
\end{equation}
In logs, the close-to-close move is the sum of the overnight move and the move after the open. Each part is made
abnormal by removing the market and the stock's sector, $\AR^{k}_{i,t} = r^{k}_{i,t} - \beta^{m}_{i}\,
r^{k}_{m,t} - \beta^{s}_{i}\,(r^{k}_{s,t} - r^{k}_{m,t})$, where $r_{m}$ is the return of the S\&P 500 ETF
(SPY), $r_{s}$ the equal-weighted return of the other members of the stock's sector, and the betas come from
the 252 sessions up to $t-1$. It is then divided by $\sigma^{k}_{i,t-1}$, a robust scale of the stock's abnormal
returns over the 126 sessions up to $t-1$. A value of $z^{k}_{i,t}$ is therefore a move in units of the stock's
typical daily move, scaled by information known before the session. Equation~\eqref{eq:split} defines what
``priced at the open'' means in this paper: news is priced at the open when a model's reading predicts $z^{on}$
but not $z^{id}$. Where the indices are clear we write $z_{cc}$, $z_{on}$ and $z_{id}$, as the tables and figures
do. Multi-day targets add up abnormal returns over their window and divide by the window's ex-ante volatility;
the main one, $y_2$, runs from the next open to the close of $t+5$.

\subsection{News, events and decision points}\label{sec:news}

The news comes from NewsWitch, a commercial company-news feed (titles and snippets with publication times). An
article is
available at its time stamp, or at the end of the hour or day when the stamp is rounded to the hour or carries
only a date. We drop items marked as non-news, price commentary and promotion, automatic institutional-holdings
reports, non-English articles, near-duplicates, and two kinds of item whose time stamp precedes its content:
explainers of a stock's move and running market stories updated in place. We fix three decision points per
session $t$:
\begin{itemize}[leftmargin=*, itemsep=1pt]
\item \textbf{D0}, at 09:15~ET, before the open. The model reads the news published since the previous close;
  a trade enters at the official open.
\item \textbf{D1}, at 10:00~ET, after the first 30-minute bar. The model also sees the move so far; a trade
  enters at the price at the end of that bar.
\item \textbf{D2}, after the close. The model reads the day's news and the day's move; a trade enters at the
  next open.
\end{itemize}
The observed reaction $G$ is the move up to the decision point in typical-move units: at D2 it is the day's
$z_{cc}$, and at D1 the abnormal move from the previous close to 10:00, divided by its ex-ante volatility.

A learned importance filter, a logistic regression on sentence embeddings of the news \citep{xiao2024}, selects
the company-days that enter the analysis. Chosen on val-A by a pre-registered rule, it captures \res{f5-recall}
of the company-days with a material price reaction at a budget of 30\% of covered company-days. An \emph{event}
is a company-session with selected news: D0 events use the news up to 09:15, D2 events the news up to the close.
The test period has \res{test-d0-events} D0 events on \res{test-d0-sessions} sessions and \res{test-d2-events}
D2 events.

Each event becomes a \emph{state}, a JSON document with fixed keys: the company's context (sector, size tier,
market sensitivity, volatility regime, recent abnormal performance, analyst-estimate revisions and whether an
earnings report is scheduled), the market's context (the previous session's market move, the level and change
of the VIX, rates, credit spreads, the dollar, oil and scheduled macroeconomic releases), attention (the number
of articles against the stock's usual number) and up to six news items with their source type and timing. Every
number passes through a codebook of words: a move of between one and two typical daily moves is ``moderate'',
followed by its multiple of a typical day's move. A post-reaction state adds the reaction so far: the stock's
move, the market's move over the same period and the volume against normal.

\paragraph{The two readings.} The \emph{pre-open reading} asks, among other questions, ``Once investors fully
absorb the news, how will the company's stock perform relative to the overall market and its sector over the
next five trading days?'' on nine levels, from a very large fall (more than three typical daily moves) to a very
large rise. With $p_k$ the probability of level $k$ and $m_k$ the level's midpoint in typical daily moves, the
\emph{news-implied move} is the model's expected move,
\[
\Nhat = \sum_{k=1}^{9} p_k m_k, \qquad m = (-3.5,\ -2.5,\ -1.5,\ -0.75,\ 0,\ 0.75,\ 1.5,\ 2.5,\ 3.5).
\] The \emph{news-implied value} $V$ maps $\Nhat$ to the realized five-day level move
with an isotonic regression \citep{barlow1972} fitted on val-A for each model: it is the model's estimate of where
the abnormal price settles once the news is absorbed. The \emph{post-reaction reading} also asks whether the
reaction so far is in the wrong direction, much too small, too small, about right, too large or much too large for
what the news justifies, and whether the move will reverse, stall or continue over the next five days. Its
judgment $J = P(\text{much too small}) + P(\text{too small}) - P(\text{too large}) - P(\text{much too large})$ is
positive when the model calls the reaction an underreaction. A pre-registered check found that the post-reaction
readings anchor on the move the model is shown, so $\Nhat$ always comes from the pre-open reading.

\paragraph{Masking.} The states hide the company. Names, aliases, tickers and cashtags become ``the company'';
person names become ``an executive'' or ``a person''; dates, weekdays, years and fiscal-period labels become
``recently'' or ``the latest quarter''; share-price levels and every percent price move of any security (the
company's, other companies', indexes, commodities and currencies) are removed. Fundamental figures, such as
earnings per share, revenue and guidance, are kept, because they are the news. An independent leak scanner,
written without the masker's rules, checked the result; the residues that remain, below half a percent of
states for names, tickers and dates, are disclosed in the deviations log. Masking cannot hide a company whose news
is distinctive, and Section~\ref{sec:memorization} measures how often the models still recognize it.

\subsection{The models}\label{sec:models}

\paragraph{Laya-Fin, the primary instrument.} Laya is an open-weights decision model built on the
ModernBERT-large encoder \citep{convaiinnovations2026,warner2025}. We fine-tuned it on the training period's states
with three heads: the nine-level move (labeled with the realized five-day move), a five-level magnitude head (the
size of the same-day move) and, for the post-reaction reading, a three-class continuation head. It was trained
once (one seed, one epoch) with AdamW \citep{loshchilov2019} and early stopping on val-A, and one temperature per
head was fitted on val-A; the pre-registered second and third seeds are robustness runs
(Section~\ref{sec:limits}).

\paragraph{Commercial decision models.} Every eligible decision model in a public model router's catalog, plus
the local Laya models, entered a two-stage zero-shot tournament on val-A. The six with the highest first-stage IC
then faced four pre-registered screens: company identification on masked states, the share of counterfactual
edits (beat or miss, raise or cut, upgrade or downgrade) that move the reading the right way, the IC gain from
unmasking, and a shuffled-news control. No model passed: all six recognized the masked company far more often than
the screen allowed (Section~\ref{sec:memorization}), and five also failed the counterfactual screen. By a decision
logged before the freeze, the three with the highest second-stage IC, d1-ultra, span-02 and drex-v2.0, were
carried to the test as \emph{screen-failing comparators} (the API models of the tables and figures). We refer to
them by short labels because the results
concern the versions we tested; the replication package lists their catalog identifiers and release dates. All
three were released in October 2026, after the sample ended.

\paragraph{The untuned model, the anchor and the baselines.} Laya zero-shot is the untuned Laya model answering
the same questions. The anchor is the December 2024 ModernBERT-large checkpoint \citep{warner2025}, fine-tuned
with the same heads and recipe as Laya-Fin; it was meant to show what training adds to a model whose
pretraining ended before the sample. After one epoch its outputs are nearly constant, so we report it as
\emph{uninformative}, not as a null, and make no claim about the effect of training. The clean baselines are a
ridge regression \citep{hoerl1970} on sentence embeddings of the news, a gradient-boosted model \citep{ke2017} on
price, volume and size alone, and a generative LLM prompted for the same nine-level forecast on a 2,000-event
subsample. A pre-registered rule chose the LLM among models whose stated knowledge cutoff (April 2024) precedes
the sample, so unlike the decision models it cannot have seen the sample in training. Two more baselines, vendor
sentiment and a gradient-boosted model on the vendor's topic tags, are built on labels the vendor produced after
the fact; they are an upper bound that may carry look-ahead, not a benchmark.

\subsection{Pre-registration, freeze and blinding}\label{sec:prereg}

The pre-registration was written on 2026-10-24, before any panel, target or model output existed, and fixes the
universe, the splits, the targets, the masking, the questions, every selection rule, the hypotheses with their
families and corrections, the strategies and the inference. Three amendments were added the same day, also before any
panel, target or model output existed: convergence to the news-implied value, a wider question battery and the
firm-level explanation analysis.
Later changes went only into an append-only, dated deviations log. On 2026-10-28, before any 2026 model call, a
SHA-256 hash over the pre-registration, every configuration file and the analysis code was recorded and tagged in
version control. The lock on the 2026 prices checks that hash. After the freeze it was opened to build the
post-reaction states (reactions only) and then to release the test outcomes to the frozen analyses, the hypothesis
tests first and the strategies later the same day; every read is logged.

\subsection{Inference}\label{sec:inference}

The main statistic is the daily information coefficient,
\begin{equation}\label{eq:ic}
\IC_t(\Nhat, z) = \rho_{S}\bigl(\{\Nhat_{i,t}\}_{i \in \mathcal{E}_t},\, \{z_{i,t}\}_{i \in \mathcal{E}_t}\bigr),
\qquad
\overline{\IC} = \frac{1}{T}\sum_{t=1}^{T} \IC_t(\Nhat, z),
\end{equation}
the Spearman rank correlation $\rho_{S}$ between the readings and the moves across the events $\mathcal{E}_t$ of
session $t$ (sessions with at least five events), averaged over the $T$ sessions of a period. Its $t$-statistic
uses Newey--West standard errors with five lags \citep{newey1987}. A positive IC means that the readings rank
the day's moves in the right order on average; the IC is the usual measure of a cross-sectional signal. The
hypotheses that news is priced at the open compare the IC with the overnight move and with the move after the
open:
\begin{equation}\label{eq:h1}
\text{H1: } \Delta = \frac{1}{T}\sum_{t}\Bigl[\IC_t(\Nhat, z^{on}) - \IC_t(\Nhat, z^{id})\Bigr] > 0,
\qquad
\text{H1b: } \frac{1}{T}\sum_{t} \IC_t(\Nhat, z^{id}) > 0.
\end{equation}
H1 says that the news is priced mostly at the open, H1b that it is not fully priced there, so that a position
taken at the open can still earn. Both are tested one-sided with a stationary block bootstrap over sessions
\citep{politis1994}, with blocks of five sessions and 10,000 draws. H1b was added to the pre-registration before
the freeze, because the pre-registered strategy that enters at the open needs it.

The five primary hypotheses (H1, H1b, H2, H3 and H4, defined where they are tested) form one family, corrected by
Holm's method at 0.05 \citep{holm1979}; the secondary hypotheses form families corrected by
Benjamini--Hochberg at $q = 0.10$ \citep{benjamini1995}. Regressions are pooled OLS with day-by-tier fixed
effects and standard errors clustered by date and by firm \citep{cameron2011}, with Driscoll--Kraay standard
errors \citep{driscoll1998} as a check. A primary result counts as a general claim only if it holds with the same
sign and $p < 0.05$ for the primary instrument and for the anchor; if it holds for the primary instrument alone,
it is model-specific. Because the anchor is uninformative and the commercial models failed the screens, no
result in this paper can be more than model-specific: every result below is a statement about the models we
tested, and we do not repeat the qualifier.

\section{Clean validation did not survive the blinded test}\label{sec:validation}

\begin{figure}[tbp]
\centering
\includegraphics{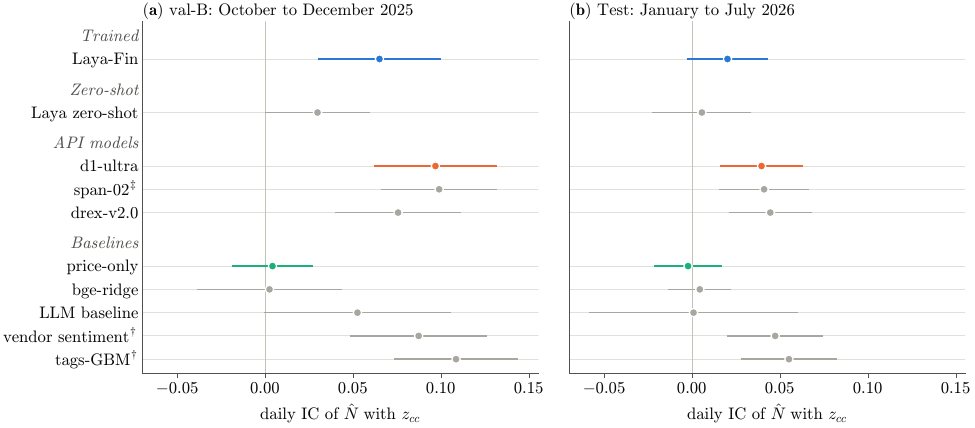}
\caption{\textbf{Every validation IC that was clearly positive fell on the blinded test: Laya-Fin's from
\res{lf-ic-valb} to \res{lf-ic-test}, d1-ultra's from \res{d1-ic-valb} to \res{d1-ic-test}.} Daily IC of the
pre-open reading $\Nhat$ with the close-to-close move $z_{cc}$ on (a) val-B (October--December 2025, the
final-selection quarter) and (b) the test period (January--July 2026), with 95\% intervals ($\pm 1.96$
Newey--West standard errors); the panels share one scale. Colors mark the series that recur across the figures:
Laya-Fin (blue), d1-ultra (orange) and the price-only benchmark (green). The LLM baseline was run on a
2,000-event subsample.
$^{\dagger}$Vendor-label baselines: the labels were produced after the fact, so their ICs are a possibly
look-ahead-contaminated upper bound. $^{\ddagger}$See Appendix~\ref{app:prereg}.}\label{fig:validation-test}
\end{figure}

Figure~\ref{fig:validation-test} compares each model's daily IC with the close-to-close move on val-B, the
quarter used for the final selection, with its IC on the test period. Laya-Fin was selected as the primary
instrument because its val-B IC, \res{lf-ic-valb} ($t = \res{lf-ic-valb-t}$), was the highest among the
candidates eligible for that role; on val-A, which it was not trained on, its IC had been \res{lf-ic-vala}. On the
test it fell to \res{lf-ic-test} ($t = \res{lf-ic-test-t}$), below the lower end of its val-B 95\% interval
(\res{lf-ic-valb-lo}). The fall is not specific to the fine-tuned model. Every model whose val-B IC was clearly
positive fell: the commercial models' ICs roughly halved (d1-ultra's from \res{d1-ic-valb} to \res{d1-ic-test}),
and the
generative LLM's fell from \res{llm-ic-valb} to \res{llm-ic-test} on its subsample. The clean price-only and
embedding benchmarks are near zero in both periods. The highest test ICs belong to the two baselines built on
vendor labels (\res{tags-ic-test} for the tag model); because the vendor produced those labels after the fact, we
read them as an upper bound that may contain look-ahead, not as a benchmark. Appendix
Table~\ref{tab:a-model-comparison} lists every model. A validation that looked clean on two quarters was weak
evidence of what the model would do next; an exploratory split (Section~\ref{sec:exploratory}) locates the decline
away from earnings days.

Table~\ref{tab:primary} reports the five primary hypotheses. None is supported for Laya-Fin (Holm-adjusted $p$
from \res{lf-h3-holm} to \res{lf-h1-holm}). H1 and H1b, whether the news is priced at the open, are the subject of
Section~\ref{sec:priced} and Table~\ref{tab:ic-target}, and H2, the over/under regression, of
Section~\ref{sec:overunder} and Table~\ref{tab:over-under}. H3 asks whether the
post-reaction judgment $J$ adds to price-only and news-only features in predicting $y_2$, the move from the next
open to the close of $t+5$; the gain in IC is \res{lf-h3} ($p = \res{lf-h3-p}$). H4 asks whether the readings of
materiality, surprise and expected magnitude predict the size of the day's move, $|z_{cc}|$, beyond a
heterogeneous autoregressive model of realized volatility \citep{corsi2009}, an earnings-day flag and the size
tier; they add \res{lf-h4} to the out-of-sample $R^2$ ($p = \res{lf-h4-p}$). Laya-Fin's magnitude head ranks the
size of the move (an AUC of \res{lf-auc} for moves of at least one typical day), so the models know which stocks
will move a lot, but no better than the stocks' own recent volatility.

\begin{table}[htbp]
\centering
\small
\begin{threeparttable}
\caption{Primary hypotheses on the test period: none of the five is supported for the primary instrument, Laya-Fin (Holm $p$ from 0.269 to 0.853).}\label{tab:primary}
\begin{tabular}{@{}lS[table-format=-1.4]cS[table-format=1.3]S[table-format=1.3]S[table-format=-1.4]S[table-format=1.3]l@{}}
\toprule
 & \multicolumn{4}{c}{Laya-Fin (primary)} & \multicolumn{2}{c}{Anchor} & \\
\cmidrule(lr){2-5}\cmidrule(lr){6-7}
Hypothesis & {Estimate} & 95\% CI & {$p$} & {Holm $p$} & {Estimate} & {$p$} & Verdict \\
\midrule
H1 & 0.005 & [$-0.035$, $0.041$] & 0.409 & 0.853 & 0.004 & 0.406 & not supported \\
H1b & 0.010 & [$-0.014$, $0.037$] & 0.213 & 0.853 & 0.002 & 0.427 & not supported \\
H2 & -0.044 & --- & 0.616 & 0.853 & 0.015 & 0.294 & not supported \\
H3 & 0.0038 & [$-0.0008$, $0.0085$] & 0.054 & 0.269 & 0.0013 & 0.017 & not supported \\
H4 & 0.0012 & [$-0.0017$, $0.0042$] & 0.216 & 0.853 & -0.0004 & 0.773 & not supported \\
\bottomrule
\end{tabular}
\begin{tablenotes}[flushleft]\footnotesize
\item The tests: H1, $\mathrm{IC}(\hat N, z_{on}) - \mathrm{IC}(\hat N, z_{id}) > 0$; H1b, $\mathrm{IC}(\hat N, z_{id}) > 0$; H2, $b > 0$ and $c < 0$ at D2; H3, $\Delta\mathrm{IC}$ of $J$ for $y_2 > 0$; H4, $\Delta R^2$ of the news for $|z_{cc}| > 0$.
\item Test period 2026-01-02 to 2026-07-31, events outside the embargo. Estimates in each hypothesis' direction; $p$ one-sided; 95\% CIs and $p$ from a stationary block bootstrap over sessions (block 5, 10,000 draws); Holm over H1, H1b, H2, H3 and H4 at 0.05.
\item H2: the estimate is $\min(b, -c)$ and $p = \max(p_b, p_c)$ (both must hold), so it has no single interval (---); $b$ and $c$ with their standard errors are in Table~\ref{tab:over-under}. Estimates and CIs to three decimals, H3's and H4's to four, since those increments are an order smaller.
\item Verdict (the pre-registered headline rule): general if Laya-Fin's Holm $p < 0.05$ with the predicted sign and the anchor's own $p < 0.05$ with the same sign; model-specific if only Laya-Fin's holds; otherwise not supported. The anchor's readings are uninformative (near-constant outputs after one epoch); its rows are reported, not read as nulls. Its $p$ is its own, not Holm-adjusted.
\item Significance: no stars are used; $p$ is one-sided unless stated.
\end{tablenotes}
\end{threeparttable}
\end{table}

\section{Pre-open news is priced at the open}\label{sec:priced}

\begin{figure}[tbp]
\centering
\includegraphics{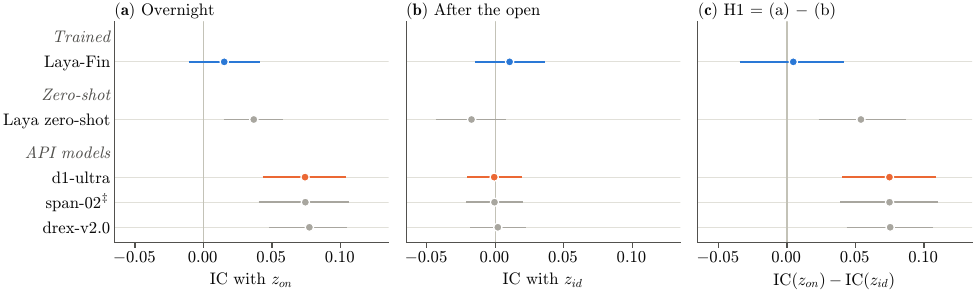}
\caption{\textbf{The commercial models' pre-open readings line up with the overnight move (daily IC
\res{d1-ic-on-test} to \res{drex-ic-on-test}), not with the move after the open (about zero).} Daily IC of the
pre-open reading $\Nhat$ on the test period with (a) the overnight move $z_{on}$ (previous close to open) and
(b) the move after the open $z_{id}$ (open to close), with 95\% bootstrap intervals; (c) H1, the difference of the
two, with its bootstrap interval. The panels share one scale. Laya-Fin's ICs are small on both sides
(\res{lf-ic-on-test} and \res{lf-ic-id-test}), so H1 and H1b are not supported for it (Holm
$p = \res{lf-h1-holm}$). $^{\ddagger}$See Appendix~\ref{app:prereg}.}\label{fig:priced-open}
\end{figure}

Equation~\eqref{eq:split} splits each day's move at the open. Figure~\ref{fig:priced-open} and
Table~\ref{tab:ic-target} show on which side of the open the models' readings predict it. For the three
commercial models the answer is clear. Their daily IC with the overnight move $z_{on}$ is \res{d1-ic-on-test} to
\res{drex-ic-on-test} ($t$ from \res{span-ic-on-test-t} to \res{drex-ic-on-test-t}), and their IC with the move
after the open, $z_{id}$, is about zero (\res{d1-ic-id-test} to \res{drex-ic-id-test}). H1, the difference, is
\res{d1-h1} for d1-ultra (95\% interval \res{d1-h1-lo} to \res{d1-h1-hi}, $p = \res{d1-h1-p}$) and about the same
for the other two. Sorted by d1-ultra's reading, the top quintile of a day's events moves \res{d1-qspread-on}
typical daily moves more overnight than the bottom quintile ($t = \res{d1-qspread-on-t}$). The untuned Laya model,
which barely recognizes masked companies (Section~\ref{sec:memorization}), shows the same pattern at a smaller
scale (H1 $= \res{lz-h1}$, $p = \res{lz-h1-p}$).

No model predicts the move after the open. H1b, the IC with $z_{id}$, is not significant for any model on the
test: \res{lf-ic-id-test} for Laya-Fin ($t = \res{lf-ic-id-test-t}$), and between \res{lz-ic-id-test} and
\res{drex-ic-id-test} for the others. This matters for Laya-Fin, whose validation signal seemed to reach past
the open: on val-B its IC with $z_{id}$ was \res{lf-ic-id-valb} ($t = \res{lf-ic-id-valb-t}$), as large as its
IC with $z_{on}$. That is why H1b was added to the pre-registration before the freeze. On the test, both of
Laya-Fin's ICs are small, and neither hypothesis is supported.

\begin{table}[htbp]
\centering
\small
\begin{threeparttable}
\caption{Daily IC by target: the API models' IC is 0.074 to 0.077 with the overnight move and about zero with the move after the open; no model predicts the move after the open.}\label{tab:ic-target}
\begin{tabular}{@{}lS[table-format=-1.3, table-space-text-pre=(, table-space-text-post=)]S[table-format=-1.3, table-space-text-pre=(, table-space-text-post=)]S[table-format=-1.3, table-space-text-pre=(, table-space-text-post=)]S[table-format=1.3, table-space-text-pre=(, table-space-text-post=)]S[table-format=1.3, table-space-text-pre=(, table-space-text-post=)]S[table-format=-1.3, table-space-text-pre=(, table-space-text-post=)]@{}}
\toprule
 & \multicolumn{3}{c}{Val-B} & \multicolumn{3}{c}{Test} \\
\cmidrule(lr){2-4}\cmidrule(lr){5-7}
Model & {$z_{cc}$} & {$z_{on}$} & {$z_{id}$} & {$z_{cc}$} & {$z_{on}$} & {$z_{id}$} \\
\midrule
\multicolumn{7}{@{}l}{\textit{Trained and zero-shot}} \\
Laya-Fin & 0.065 & 0.045 & 0.054 & 0.020 & 0.015 & 0.010 \\
 & {(}3.64{)} & {(}2.93{)} & {(}2.89{)} & {(}1.67{)} & {(}1.08{)} & {(}0.82{)} \\
anchor & -0.020 & -0.002 & -0.011 & 0.008 & 0.006 & 0.002 \\
 & {(}-1.52{)} & {(}-0.16{)} & {(}-0.75{)} & {(}0.83{)} & {(}0.50{)} & {(}0.18{)} \\
Laya zero-shot & 0.029 & 0.047 & 0.002 & 0.005 & 0.037 & -0.017 \\
 & {(}1.94{)} & {(}2.55{)} & {(}0.12{)} & {(}0.37{)} & {(}3.22{)} & {(}-1.25{)} \\
\addlinespace
\multicolumn{7}{@{}l}{\textit{API models (model-specific)}} \\
d1-ultra & 0.097 & {---} & {---} & 0.039 & 0.074 & -0.001 \\
 & {(}5.38{)} & {---} & {---} & {(}3.23{)} & {(}4.74{)} & {(}-0.07{)} \\
span-02$^{\ddagger}$ & 0.099 & {---} & {---} & 0.041 & 0.074 & -0.001 \\
 & {(}5.83{)} & {---} & {---} & {(}3.10{)} & {(}4.49{)} & {(}-0.05{)} \\
drex-v2.0 & 0.075 & {---} & {---} & 0.044 & 0.077 & 0.002 \\
 & {(}4.10{)} & {---} & {---} & {(}3.70{)} & {(}5.34{)} & {(}0.18{)} \\
\bottomrule
\end{tabular}
\begin{tablenotes}[flushleft]\footnotesize
\item Mean daily Spearman IC of the pre-open news reading $\hat N$ with the close-to-close move $z_{cc}$, the overnight move $z_{on}$ (previous close to open) and the move after the open $z_{id}$ (open to close), each in typical daily moves, to three decimals; Newey--West $t$ in parentheses, to two; D0 events outside the embargo (val-B 2025-10-01 to 2025-12-31; test 2026-01-02 to 2026-07-31).
\item Val-B ICs with $z_{on}$ and $z_{id}$ come from a run of the frozen specification on val-B before the freeze, which covered the trained and zero-shot Laya models only (---: not computed).
\item The anchor's readings are uninformative (near-constant outputs after one epoch); its rows are reported, not read as nulls. The API models fail the pre-registered contamination screens, so their results are model-specific. $^{\ddagger}$span-02: after the freeze, a stability check that repeats earlier requests (the canary) fired for it; its build was unchanged, and production went ahead under a rule recorded in the deviations log.
\end{tablenotes}
\end{threeparttable}
\end{table}

\paragraph{What the overnight IC measures.} The overnight move includes trading after the previous close and
before the open, and pre-market trading starts at 04:00~ET. Part of $z_{on}$ has therefore already happened when
the model reads the news at 09:15, so the overnight IC partly measures how well a reading lines up with a move
the market is already making, not only a forecast of the move between 09:15 and the open. The pre-registered
analysis scores the readings against the whole overnight move and does not split it at 09:15. Either way, the
part of the news that the readings capture is in the price by the open, and nothing measurable is left for a
trader who enters there.

\paragraph{Robustness.} No pre-registered robustness variant changes these conclusions
(Appendix Figure~\ref{fig:a-robustness}): dollar-volume tiers, dropping articles with date-only or
hour-rounded time stamps, a non-overlapping sample, placebo-adjusted targets, a 60-session volatility, dropping
delisting horizons and dropping text about price moves. Unmasking does not raise d1-ultra's IC: on
\res{unmask-n} test events its IC with $z_{cc}$ is \res{unmask-ic-masked} with masked states and
\res{unmask-ic-unmasked} with the same states unmasked. The timing of the news matters. d1-ultra's overnight IC
is \res{d1-ic-on-0815} ($t = \res{d1-ic-on-0815-t}$) when the latest article arrives between 08:00 and 09:15,
\res{d1-ic-on-0408} when it arrives between 04:00 and 08:00, and not significant when it precedes 04:00
(\res{d1-ic-on-pre04}). The closer to the open the news arrives, the more of the overnight move the reading
lines up with.

\section{The models cannot tell over- from under-reaction}\label{sec:overunder}

\subsection{The over/under regression}\label{sec:h2}

The central pre-registered hypothesis, H2, asks whether a model's reading of the news separates the part of a
day's move that will persist from the part that will reverse. After the close (D2) we regress the next week's
move on the news-implied move and on the day's move:
\begin{equation}\label{eq:h2}
y_{i,t} = a + b\,\Nhat_{i,t} + c\,G_{i,t} + \alpha_{d(t)\times\tau(i)} + \varepsilon_{i,t},
\end{equation}
where $y_{i,t} = y_2$ is the abnormal move from the next open to the close of $t+5$ in typical-move units,
$\Nhat_{i,t}$ is the model's reading of the day's news (all of it, up to the close, with price language removed),
$G_{i,t} = z^{cc}_{i,t}$ is the day's move and $\alpha_{d(t)\times\tau(i)}$ is a day-by-tier fixed effect; there is
no conditioning on the size of $G$. H2 predicts $b > 0$ (moves the news explains persist: underreaction to news)
and $c < 0$ (moves the news does not explain reverse: overreaction to price); both must hold, and each is tested
one-sided with standard errors clustered by date and firm.

Table~\ref{tab:over-under} reports the estimates. For Laya-Fin, $b = \res{lf-h2-b-d2}$ ($p = \res{lf-h2-bp-d2}$)
and $c = \res{lf-h2-c-d2}$ ($p = \res{lf-h2-cp-d2}$) on \res{lf-h2-n-d2} events. The reversal coefficient has the
predicted sign for every model but is never significant, and the news coefficient is not positive: d1-ultra's is
\res{d1-h2-b-d2} ($p = \res{d1-h2-bp-d2}$ for a positive effect). The two earlier decision points, reported
outside the family, do not change the picture. None of the eight pre-registered heterogeneity cuts (the three size
tiers, quantified against story news, small moves, press wires, low usual coverage and busy days) changes
Laya-Fin's coefficients (Benjamini--Hochberg $q \geq \res{het-q-min}$; Appendix Table~\ref{tab:a-heterogeneity}),
and neither does the style of the coverage: a more promotional or sensational tone, a dimension of text separate
from its content \citep{wegmann2026}, does not predict reversal (H6; Appendix Table~\ref{tab:a-secondary}).

\begin{table}[htbp]
\centering
\small
\begin{threeparttable}
\caption{The over/under regression (H2): for Laya-Fin the news coefficient $b$ is $-$0.044 and the reversal coefficient $c$ is $-$0.015 ($p$ = 0.110): $c$ has the predicted sign but is not significant, and $b$ is not positive at the close.}\label{tab:over-under}
\begin{tabular}{@{}lS[table-format=-1.3]S[table-format=1.3, table-space-text-pre=(, table-space-text-post=)]S[table-format=1.3]S[table-format=-1.3]S[table-format=1.3, table-space-text-pre=(, table-space-text-post=)]S[table-format=1.3]S[table-format=5.0]@{}}
\toprule
Decision point & {$b$} & {SE} & {$p_b$} & {$c$} & {SE} & {$p_c$} & {Events} \\
\midrule
\multicolumn{8}{@{}l}{\textit{Laya-Fin (primary)}} \\
D2, close (H2) & -0.044 & {(}0.149{)} & 0.616 & -0.015 & {(}0.012{)} & 0.110 & 17188 \\
D1, 10:00 & 0.229 & {(}0.148{)} & 0.062 & 0.013 & {(}0.012{)} & 0.861 & 13203 \\
D0, open & 0.278 & {(}0.212{)} & 0.096 & {---} & {---} & {---} & 13439 \\
\addlinespace
\multicolumn{8}{@{}l}{\textit{Anchor (uninformative)}} \\
D2, close (H2) & 0.748 & {(}1.377{)} & 0.294 & -0.015 & {(}0.012{)} & 0.111 & 17188 \\
D1, 10:00 & 1.964 & {(}2.482{)} & 0.215 & 0.014 & {(}0.012{)} & 0.874 & 13203 \\
D0, open & -4.510 & {(}3.236{)} & 0.917 & {---} & {---} & {---} & 13439 \\
\addlinespace
\multicolumn{8}{@{}l}{\textit{d1-ultra (model-specific)}} \\
D2, close (H2) & -0.076 & {(}0.031{)} & 0.991 & -0.011 & {(}0.012{)} & 0.171 & 17188 \\
D1, 10:00 & -0.018 & {(}0.032{)} & 0.711 & 0.015 & {(}0.012{)} & 0.884 & 13203 \\
D0, open & 0.022 & {(}0.037{)} & 0.277 & {---} & {---} & {---} & 13439 \\
\bottomrule
\end{tabular}
\begin{tablenotes}[flushleft]\footnotesize
\item $y = a + b\,\hat N + c\,G$ with day $\times$ tier fixed effects, on the test period: at D2 $y = y_2$ (next open to $t{+}5$), $\hat N$ from the day's news and $G$ the day's move; at D1 the move after 10:00 on $\hat N$ from the pre-open news and the move to 10:00; at D0 $y = z_{id}$ and there is no $G$, so no $c$ (---). Two-way clustered (session, firm) standard errors in parentheses; $p_b$ tests $b > 0$ and $p_c$ tests $c < 0$, one-sided. H2 is the D2 row of Laya-Fin.
\item The anchor's readings are uninformative (near-constant outputs after one epoch); its rows are reported, not read as nulls. The API models fail the pre-registered contamination screens, so their results are model-specific.
\item Significance: no stars are used; $p$ is one-sided unless stated.
\end{tablenotes}
\end{threeparttable}
\end{table}

\subsection{Convergence to the news-implied value}\label{sec:convergence}

The convergence analysis asks the same question along the whole path from the open to $t+20$. For each rung
$h$ of a ladder from the open to the close of $t+20$, we regress the level move from the previous close to $h$
on the news-implied value:
\begin{equation}\label{eq:ladder}
\CARlvl_{i,h} = \kappa_h V_i + \alpha_{d(t)\times\tau(i)} + \varepsilon_{i,h},
\qquad h \in \{O_t,\ \text{10:00},\ C_t,\ C_{t+1},\ C_{t+2},\ C_{t+5},\ C_{t+10},\ C_{t+20}\},
\end{equation}
where $\CARlvl_{i,h}$ is the cumulative abnormal return in typical daily moves, without square-root-of-time
scaling, so that it is comparable with $V$. $\kappa_h = 1$ means that by rung $h$ the price has moved all the way
to the model's value; a rising $\kappa_h$ means drift toward it and a falling one means reversal.

\begin{figure}[tbp]
\centering
\includegraphics{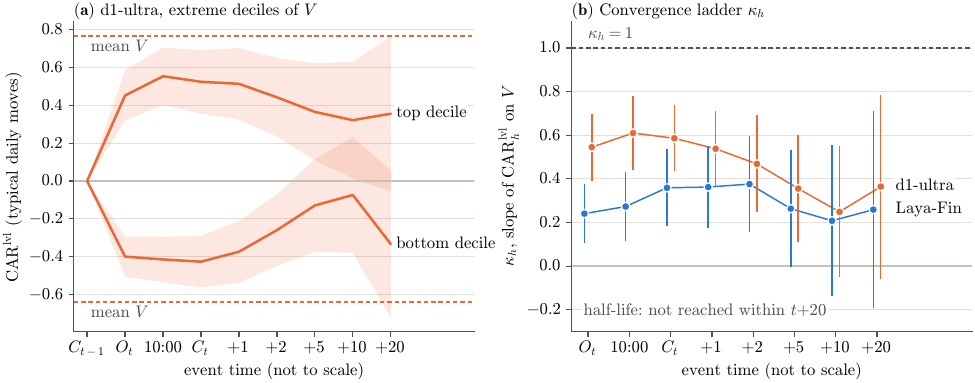}
\caption{\textbf{Prices move part of the way to the model's news-implied value by the close, then partly fade:
the convergence slope is \res{lf-kappa-ct} at the close for Laya-Fin and \res{d1-kappa-ct} for d1-ultra, and the
gap's half-life is not reached within 20 sessions.} Test period. Event time is not to scale: $O_t$ is the open,
10:00 the D1 decision point, $C_t$ the close of day $t$ and $+k$ the close of session $t+k$. (a) d1-ultra's top
and bottom deciles of the news-implied value $V$: the level move $\CARlvl$ from the previous close, in typical
daily moves, with $\pm 1.96$ event-level standard-error bands (descriptive), against each decile's mean $V$
(dashed). (b) The convergence ladder $\kappa_h$ of Equation~\eqref{eq:ladder}, with 95\% intervals from
standard errors clustered by date and firm, against full convergence at $\kappa_h = 1$.}\label{fig:convergence}
\end{figure}

Prices move part of the way to the model's value on day $t$ and no further (Figure~\ref{fig:convergence};
Appendix Table~\ref{tab:convergence}). For Laya-Fin, $\kappa_h$ is \res{lf-kappa-ct} at the close and
\res{lf-kappa-t20} at $t+20$; for d1-ultra the path is higher and has the same shape, from \res{d1-kappa-ct} at
the close to \res{d1-kappa-t20} at $t+20$. Neither pre-registered increase from the open to $t+20$ is significant
($p = \res{lf-kappa-incr-p}$ for Laya-Fin and \res{d1-kappa-incr-p} for d1-ultra). Prices also drift in $V$'s
direction before the event: five sessions earlier, d1-ultra's top decile stood \res{d1-dec10-pre5} and its bottom
decile \res{d1-dec1-pre5} relative to the previous close. The states describe the company's recent abnormal
performance in words, so part of $V$ reflects momentum.

Because $V$ is a model's estimate, a slope below one does not by itself mean that the market underreacts. Under
classical measurement error, the level of $\kappa_h$ mixes market efficiency with model noise, while the shape of
the ladder (a rise or a fall after the decision point) is identified up to a positive factor
(Appendix~\ref{app:convergence}). The informative fact is therefore the fade after the close: the part of the
day's move that lines up with the news does not continue, and partly gives back, although the pre-registered
increase is not significant, so the fade is descriptive. The gap between the model's value and the price does not
close either: its half-life is not reached within 20 trading days for any model, and because $V$ is small next to
the day's move, the pre-registered gap-closure test measures how much of the day's move reverses, not convergence
to $V$ (Appendix~\ref{app:convergence}).

\subsection{Under, over and the size of the move}\label{sec:rg}

\begin{figure}[tbp]
\centering
\includegraphics{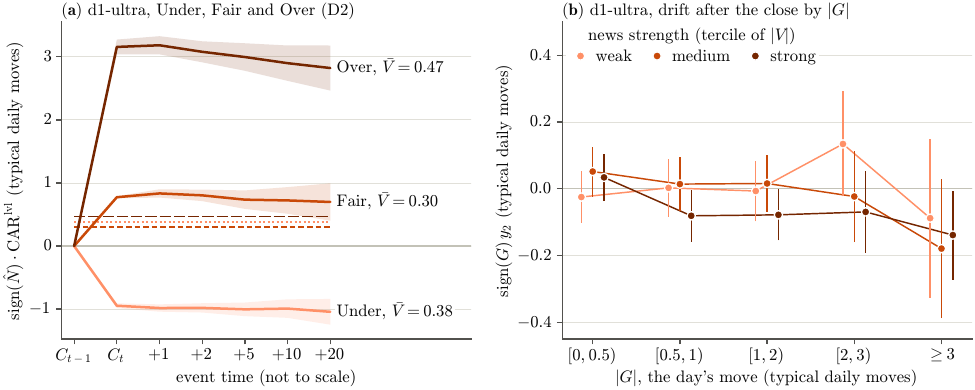}
\caption{\textbf{The models' ``underreaction'' calls never catch up, and moves of three or more typical days
reverse slightly in every tercile of news strength (\res{drift-g3-weak} to \res{drift-g3-medium} typical moves
over the next week).} Test period,
d1-ultra. (a) D2 events by the post-reaction class, pooled over the size tiers and signed in the
news' direction ($\mathrm{sign}(\Nhat)\cdot\CARlvl$), with $\pm 1.96$ event-level standard-error bands, against
each class's mean signed value $\bar V$ (dashed: dotted for Under, short dashes for Fair, long dashes for Over).
(b) Drift after the close, $\mathrm{sign}(G)\cdot y_2$ (next open to the close of $t+5$, in the direction of the
day's move), by the size of the day's move $|G|$, one line per tercile of news strength $|V|$, with 95\%
intervals; for moves of three or more, two of the three intervals include zero. The strong tercile also reverses
slightly for moves of 0.5 to 2 typical days. The pre-registered
regression (Table~\ref{tab:over-under}) finds no support for H2.}\label{fig:over-under}
\end{figure}

The post-reaction reading classifies each event as an underreaction, a fair reaction or an overreaction, by how
far the day's move fell short of or exceeded the model's value. Figure~\ref{fig:over-under}~(a) follows
d1-ultra's classes after the close, signed in the direction of the news. The \res{rg-under-n} ``Under'' events
moved against the news on day $t$ and never catch up (\res{rg-under-t20} typical moves at $t+20$, against a mean
value of \res{rg-under-v}); the ``Over'' events give back part of their move, from \res{rg-over-t0} at the close
to \res{rg-over-t20}. Laya-Fin calls too few events either way to draw. Panel (b) asks the question by the size of
the move instead: moves of at least three typical days reverse slightly in every tercile of news strength (two of
the three intervals include zero), and smaller moves show small means of both signs. There is no sign that
important news with a small move is
underpriced.

Two descriptive cuts, pre-registered for reporting but not as tests, show where convergence is strongest.
Laya-Fin's slope at the close is \res{lf-kappa-ct-top100} for the 100 largest companies against
\res{lf-kappa-ct-mid} and \res{lf-kappa-ct-rest} for the other tiers. And by d1-ultra's own label of the news,
prices at $t+20$ reflect the value of numerical news ($\kappa_{t+20} = \res{d1-kappa-t20-numbers}$,
$p = \res{d1-kappa-t20-numbers-p}$) but not that of narrative news (\res{d1-kappa-t20-narrative}), a pattern that
echoes the earlier finding that markets treat numbers and stories differently \citep{kargarzadeh2026}.

\section{News explains little of a stock's own move; the commercial models are overconfident}
\label{sec:explaining}

\subsection{Explaining realized moves}\label{sec:ladder}

\citet{roll1988} asked how much of a stock's daily return public news explains, and \citet{kelly2026} revisit
the question for the market with generative AI. We ask it for single stocks, out of sample. For every covered
company-day of the test period (days without selected news enter with zero news features, as in Roll), we build
up the explanation of the stock-specific move $z_{cc}$ step by step: the market and the sector (Roll's first
step), then the pre-open news reading, then the reading of all the day's news, and finally the post-reaction
reading's explanation of the move, which reads the realized move and is therefore an attribution, not a
forecast. Each step is a ridge regression fitted on validation data by the pre-registered protocol and scored on
the test by
\begin{equation}\label{eq:r2}
R^{2}_{\mathrm{OOS}} = 1 - \frac{\sum_{i,t}\bigl(y_{i,t} - \hat y_{i,t}\bigr)^{2}}
{\sum_{i,t}\bigl(y_{i,t} - \bar y_{\mathrm{ref}}\bigr)^{2}},
\qquad
\Delta R^{2}_{s} = R^{2}_{\mathrm{OOS}}(s) - R^{2}_{\mathrm{OOS}}(s-1),
\end{equation}
where $\bar y_{\mathrm{ref}}$ is the reference the frozen specification fixes (a zero forecast for the
stock-specific move, and the val-B mean for the size of the move in H4), $R^{2}_{\mathrm{OOS}}(s)$ is the fit of
steps one to $s$, and $\Delta R^{2}_{s}$ is what step $s$ adds to the steps before it. A positive
$R^{2}_{\mathrm{OOS}}$ means the step predicts better than the reference on data it was not fitted on.

\begin{figure}[tbp]
\centering
\includegraphics{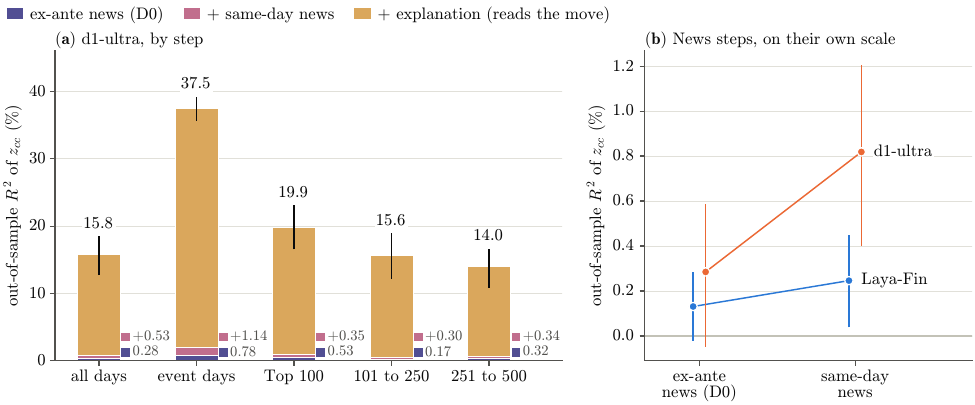}
\caption{\textbf{Same-day news explains \res{lf-ladder-post-pct} of the stock-specific daily variance out of
sample for Laya-Fin and \res{d1-ladder-post-pct} for d1-ultra; reading the realized move explains
\res{d1-ladder-expl-pct}.} Test period; out-of-sample $R^2$ in percent of the stock-specific variance (the market
and the sector removed). (a) d1-ultra's $R^2$ by cut, stacked by step in the legend's order: the pre-open news,
plus the day's news, plus the post-reaction explanation (which reads the realized move, so it is not a
forecast). Each total stands above its bar with its 95\% bootstrap interval; the two thin news steps are
labeled beside each bar. For comparison, the market and the sector explain \res{roll-r2-pct} of the total
daily variance, a different base. (b) The news steps on their own scale, with 95\% bootstrap intervals, for
Laya-Fin and d1-ultra. Only d1-ultra answers the explanation question.}\label{fig:explaining}
\end{figure}

The news explains little of a stock's own move (Figure~\ref{fig:explaining}, Table~\ref{tab:explaining}). The
market and the sector explain \res{roll-r2-pct} of the total daily variance. Of the stock-specific part, the
reading of all the day's news explains \res{lf-ladder-post-pct} for Laya-Fin and \res{d1-ladder-post-pct} for
d1-ultra, much of it already knowable before the open. H5, the secondary hypothesis that this step is positive,
is supported for both models (Benjamini--Hochberg $q = \res{lf-h5-q}$ for Laya-Fin). Reading the realized move
explains far more: d1-ultra's explanation step reaches \res{d1-ladder-expl-pct} of the stock-specific variance on
all covered days and \res{d1-ladder-expl-event-pct} on days with news. Yet being explained does not predict
persistence. In the pre-registered persistence regression the coefficient on the explanation is \res{theta-d2}
after the close ($p = \res{theta-d2-p}$ for a positive effect): explained moves do not persist more than
unexplained ones, and if anything reverse slightly more.

\begin{table}[htbp]
\centering
\small
\begin{threeparttable}
\caption{Explaining realized moves: same-day news explains 0.25\% of the stock-specific variance for Laya-Fin (H5, supported) and 0.82\% for d1-ultra; reading the move explains 15.8\%.}\label{tab:explaining}
\begin{tabular}{@{}lS[table-format=2.3]S[table-format=1.2]S[table-format=2.3, table-space-text-pre={$<$}]S[table-format=2.2]S[table-format=2.3]S[table-format=1.2]@{}}
\toprule
 & \multicolumn{2}{c}{Laya-Fin} & \multicolumn{2}{c}{d1-ultra} & \multicolumn{2}{c}{Anchor} \\
\cmidrule(lr){2-3}\cmidrule(lr){4-5}\cmidrule(lr){6-7}
Step & {$R^2$} & {$\Delta R^2$} & {$R^2$} & {$\Delta R^2$} & {$R^2$} & {$\Delta R^2$} \\
\midrule
\multicolumn{7}{@{}l}{\textit{Panel A: the explanation ladder, out-of-sample $R^2$ of $z_{cc}$ (\%)}} \\
+ ex-ante news (D0) & 0.13 & {---} & 0.28 & {---} & 0.01 & {---} \\
+ same-day news & 0.25 & 0.12 & 0.82 & 0.53 & 0.03 & 0.02 \\
+ explanation (reads the move) & {---} & {---} & 15.76 & 14.94 & {---} & {---} \\
share knowable before the open (\%) & 53.17 & {---} & 34.78 & {---} & 35.63 & {---} \\
\addlinespace
\multicolumn{7}{@{}l}{\textit{Panel B: the secondary test H5}} \\
H5: same-day news step, $R^2 > 0$ & 0.25 & {---} & 0.82 & {---} & 0.03 & {---} \\
one-sided $p$ & 0.010 & {---} & {$<$}0.001 & {---} & 0.072 & {---} \\
BH $q$ & 0.058 & {---} & {$<$}0.001 & {---} & 0.358 & {---} \\
\bottomrule
\end{tabular}
\begin{tablenotes}[flushleft]\footnotesize
\item Covered member-days of the test period; out-of-sample $R^2$ of the stock-specific move $z_{cc}$ (the market and sector removed) by cumulative step, in \%; $\Delta R^2$ is the step's increment. Roll's first step, the market and the sector, explains 25.3\% of total daily variance, a different base. The explanation step needs the post-reaction reading's explanation answer, which only d1-ultra gives; it reads the realized move, so it is not a prediction. H5 is the only secondary test that survives Benjamini--Hochberg ($q = 0.10$) for Laya-Fin. ---: not applicable (the first step has no increment; only d1-ultra answers the explanation question; the share and Panel B have one value per model). $R^2$, $\Delta R^2$ and the share in \% to two decimals; $p$ and $q$ to three.
\item The anchor's readings are uninformative (near-constant outputs after one epoch); its rows are reported, not read as nulls. The API models fail the pre-registered contamination screens, so their results are model-specific.
\end{tablenotes}
\end{threeparttable}
\end{table}

\subsection{Calibration}\label{sec:calibration}

The pre-registered calibration metric is the expected calibration error \citep{pakdamannaeini2015,guo2017}: the
pre-open reading's nine levels are grouped into three classes (a fall, no move and a rise, split at half a
typical daily move), and the top class's stated probability is compared with its hit rate against the realized
five-day move, in 15 equal-width bins. Laya-Fin's error is small (\res{lf-ece}) only because its probabilities
barely leave the validation base rates: its log-loss improves on them by \res{lf-llskill}. The commercial models
are overconfident. Their stated probabilities overstate their hit rates, with errors from \res{drex-ece} to
\res{span-ece} and negative log-loss skill, and a single temperature fitted on val-A lowers d1-ultra's error from
\res{d1-ece-before} to \res{d1-ece-after} (Appendix Table~\ref{tab:a-calibration} and
Figure~\ref{fig:a-reliability}). Calibration does not come with the decision-model interface; it has to be
checked.

\section{What the models know about masked firms}\label{sec:memorization}

Masking removes names, but news is distinctive: a drug approval, a store count or a guidance range can point to
one company. We measure how often a model recognizes the company behind a masked state. Each of \res{mem-n}
masked val-A states is shown with four candidate companies from the same sector and size tier, the true one
and three distractors chosen to match its news activity, and the model picks one. For model $m$,
\begin{equation}\label{eq:memorization}
\mathrm{Acc}_m = \frac{1}{n}\sum_{i=1}^{n} \mathbf{1}\bigl[\hat c_{i} = c_{i}\bigr],
\qquad
\text{excess}_m = \mathrm{Acc}_m - p_0,
\end{equation}
where $c_i$ is the true company, $\hat c_i$ the model's choice and $p_0 = \res{mem-baseline-f2}$ the matched
baseline: the accuracy of a guess that uses no reading, only the candidates' news activity. Intervals are
Wilson score intervals \citep{wilson1927}. The matched baseline matters: with distractors drawn at random, a
guess based on news activity alone already scores \res{mem-baseline-na06}, because the true company always has
news.

\begin{figure}[tbp]
\centering
\includegraphics{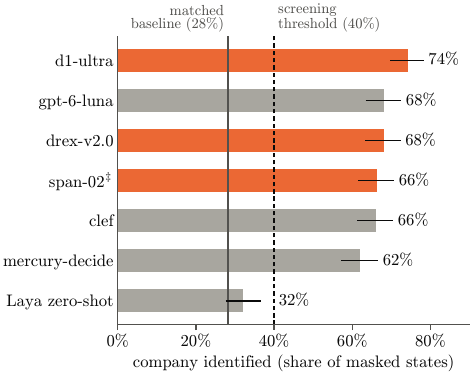}
\caption{\textbf{Commercial decision models identify the company behind masked news in \res{mem-min} to
\res{mem-max} of states, against a matched baseline of \res{mem-baseline}; the open-weights Laya model does so in
\res{mem-laya}.} Company identification on masked val-A states with activity-matched distractors: the share of
states in which each model names the company among four candidates, sorted, with its Wilson 95\% interval,
against the matched baseline (a guess that uses no reading; solid line) and the pre-registered screening
threshold (40\%; dashed line). The three commercial models carried to the test are in the comparator color,
the other commercial models and Laya zero-shot in gray. Every model was released after the sample ended.
Identifying a company is not the same as remembering its returns. $^{\ddagger}$See
Appendix~\ref{app:prereg}.}\label{fig:memorization}
\end{figure}

The commercial decision models recognize the company far more often than the baseline
(Figure~\ref{fig:memorization}): from \res{mem-min} of states (mercury-decide) to \res{mem-max} (d1-ultra), and
\res{mem-span} and \res{mem-drex} for the other two comparators. The open-weights Laya model, the base of
Laya-Fin, does so in \res{mem-laya}, close to the baseline. All of the commercial models therefore fail the
pre-registered screen of 40\%, and every model in the figure was released after the sample ended: Laya on
\res{rel-laya} and the commercial models between \res{rel-drex} and \res{rel-span}. Their training data are
undisclosed, so we cannot rule out that they saw text from the test period.

Recognition is not the same as remembering returns, and three results suggest that it did not drive the
commercial models' overnight signal. The open-weights Laya model, which recognizes masked companies barely more
often than the baseline, shows the same split between the overnight move and the move after the open
(Section~\ref{sec:priced}); unmasking the states does not raise d1-ultra's IC; and the commercial models' ICs fell
from validation to test as the other signals' did, which is not what an advantage from having seen the test
period would produce. None of the three is a test of memorization. The result is, however, a warning for
backtests of language models on historical news: masking hides names, not companies. The pre-registered forward
test (Section~\ref{sec:limits}) asks the question directly, on news published after the models were released.

\section{Economic significance}\label{sec:strategies}

The pre-registered strategies turn each reading into a daily long-short book within size tiers, hedged against
the market with SPY. The \emph{convergence strategy} (S-CONV) buys the events whose price lies furthest below the
model's value (the top 20\% of the gap $U$ between the value and the move so far) and sells those furthest above
it, after the first half hour (D1) or after the close (D2). Three more families rank the pre-open reading,
entering at the open (S-D0), and the post-reaction judgment $J$, signed by the move, at 10:00 (S-D1) and after the
close (S-D2). An RG book trades only the events the model calls an under- or over-reaction, and a gated book only
its most confident readings. Costs are charged at 5, 10 and 20 basis points per side on both legs and on the
hedge. Each book is compared with 100 books built on random signals over the same events, and its Sharpe ratio is
deflated \citep{bailey2014} for the \res{ntrials-strat} trials the ledger held when the strategies were evaluated:
every selection of filters, encodings, models and baselines, and every strategy variant up to then
(Appendix~\ref{app:dsr}).

\begin{figure}[tbp]
\centering
\includegraphics{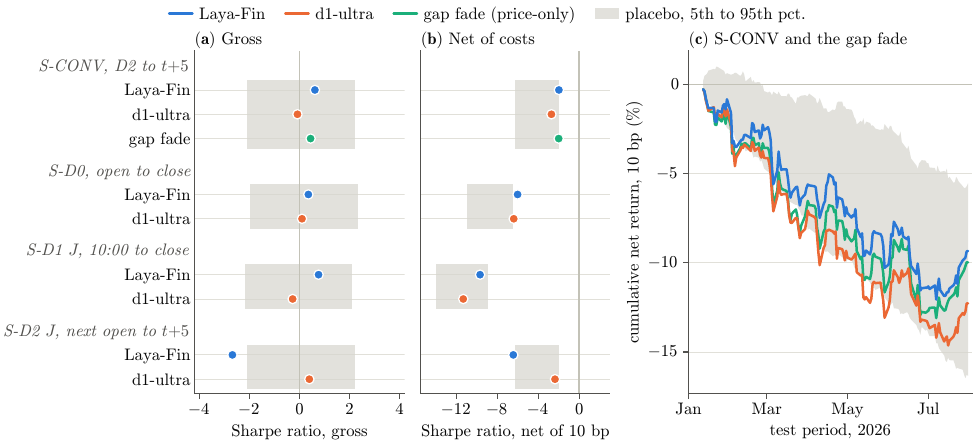}
\caption{\textbf{No pre-registered strategy survives costs: net of 10 basis points every main book loses, and
the convergence strategy is effectively a price-only gap fade (\res{sconv-shared} of positions shared, daily
correlation \res{sconv-corr}).} Test period. (a, b) Annualized Sharpe ratio of each main book, gross and net of
10 basis points per side on both legs and the SPY hedge, at the exit of its hypothesis' horizon; the gray box is
the 5th to 95th percentile of 100 random-signal books on the same events; (b) has its own scale. (c) The
convergence strategy's cumulative net return (Laya-Fin and d1-ultra) against the price-only gap fade, with the
placebo's 5th to 95th percentile band.}\label{fig:strategies}
\end{figure}

No pre-registered strategy is worth trading after costs (Table~\ref{tab:strategies} and
Figure~\ref{fig:strategies}). Gross of costs, no book reaches the 95th percentile of its random-signal placebo,
and Laya-Fin's post-reaction book after the close falls below the 5th (percentile \res{sd2j-pct}), as expected
of judgments that carry no positive information about the next week. Net of 10 basis points every main book
loses except the tiny RG book, which holds one stock a side on a handful of days. The convergence strategy is, in
effect, a price-only gap fade: because the model's value is small next to the day's move, ranking by the gap buys
the day's losers and sells its winners. The two books share \res{sconv-shared} of their positions, and their
daily returns correlate \res{sconv-corr}. The highest gross Sharpe ratio, \res{sd0-gated-gross} for the pre-open
book restricted to Laya-Fin's most confident readings and held to the next close, breaks even at
\res{sd0-gated-be} basis points per side, and the one-day books turn over about \res{sd0-turnover} times their
capital a day, so costs decide their sign. The largest deflated-Sharpe probability of any test book is
\res{dsr-max}. Short-sale borrow costs, which would lower every net figure, are not modeled.

\begin{table}[htbp]
\centering
\small
\setlength{\tabcolsep}{3pt}
\begin{threeparttable}
\caption{Strategies on the test period: no pre-registered book survives costs and the deflated Sharpe ratio (the largest deflated-Sharpe probability is 0.024).}\label{tab:strategies}
\begin{tabular}{@{}llS[table-format=-1.2]S[table-format=-2.2]S[table-format=1.2]S[table-format=-2.1]S[table-format=2.0]S[table-format=1.3, table-space-text-pre={$<$}]@{}}
\toprule
 & & \multicolumn{2}{c}{Sharpe ratio} & & & & \\
\cmidrule(lr){3-4}
Book & Holding & {Gross} & {Net} & {Turnover} & {Break-even (bp)} & {Placebo pct.} & {DSR} \\
\midrule
\multicolumn{8}{@{}l}{\textit{Laya-Fin (primary)}} \\
S-CONV & D2 to $t{+}5$ & 0.61 & -1.98 & 0.90 & 2.4 & 68 & {$<$}0.001 \\
S-D0 & open to close & 0.35 & -6.03 & 4.36 & 0.5 & 64 & {$<$}0.001 \\
S-D0, gated & open to $t{+}1$ & 1.88 & -0.38 & 2.31 & 8.3 & 93 & {$<$}0.001 \\
S-D1 $J$ & 10:00 to close & 0.76 & -9.70 & 4.36 & 0.7 & 71 & {$<$}0.001 \\
S-D2 $J$ & next open to $t{+}5$ & -2.69 & -6.45 & 0.83 & -7.1 & 3 & {$<$}0.001 \\
S-D2 RG & next open to $t{+}5$ & 1.69 & 1.50 & 0.05 & 85.1 & 92 & 0.024 \\
\addlinespace
\multicolumn{8}{@{}l}{\textit{d1-ultra (model-specific)}} \\
S-CONV & D2 to $t{+}5$ & -0.09 & -2.72 & 0.90 & -0.3 & 48 & {$<$}0.001 \\
S-D0 & open to close & 0.10 & -6.41 & 4.41 & 0.1 & 58 & {$<$}0.001 \\
S-D1 $J$ & 10:00 to close & -0.28 & -11.35 & 4.51 & -0.3 & 40 & {$<$}0.001 \\
S-D2 $J$ & next open to $t{+}5$ & 0.39 & -2.37 & 0.85 & 1.4 & 64 & {$<$}0.001 \\
\addlinespace
\multicolumn{8}{@{}l}{\textit{Benchmarks}} \\
gap fade (price-only) & D2 to $t{+}5$ & 0.44 & -2.01 & 0.90 & 1.8 & 66 & {$<$}0.001 \\
vendor sentiment$^{\dagger}$ & open to close & 0.27 & -6.24 & 4.36 & 0.4 & 62 & {$<$}0.001 \\
\bottomrule
\end{tabular}
\begin{tablenotes}[flushleft]\footnotesize
\item Annualized Sharpe ratios ($\sqrt{252}$), gross and net of 10 bp per side on both legs and the SPY hedge; no borrow cost. Turnover: dollars traded a day per dollar of capital. Break-even: the cost per side, in basis points, at which the mean net return is zero (gross mean $\div$ turnover). Placebo: the percentile of the gross Sharpe among 100 random-signal books on the same events. DSR: the deflated-Sharpe probability at 10 bp with $N$ = 545 trials, every trial logged when these pre-registered strategies were evaluated (the ledger holds 1,620 after the post hoc exploratory study). Each family is shown at its hypothesis' horizon; every book is in the replication package. S-CONV ranks by the gap $U = V - G^{\mathrm{lvl}}$; S-D0 by $\tilde N$; S-D1 and S-D2 by the judgment $J$ signed by the move; RG books trade the cases the model calls a misreaction (one name a side).
\item $^{\dagger}$Vendor sentiment: its labels were produced retrospectively, so its books are a possibly look-ahead-contaminated upper bound, not a benchmark; the clean benchmarks are the price-only gap fade and the random-signal placebo. The API models fail the pre-registered contamination screens, so their results are model-specific.
\end{tablenotes}
\end{threeparttable}
\end{table}

\section{Exploratory analyses (post hoc): where the signal lives}\label{sec:exploratory}

Everything in this section is exploratory. The analyses were requested after unblinding, they are outside every
pre-registered family and they support no claim; they suggest where to look, and the forward test of
Section~\ref{sec:limits} re-tests two of them as pre-registered hypotheses. Before this study was designed, the
authors' AI research assistant had looked at two of the statistics below on the test period (hit rates and mean
moves by conviction band, and the statistics of the 10:00 reading), and those looks shaped it. Every exploratory
variant is logged as a trial: the ledger grew from \res{ntrials-strat} to \res{ntrials-total} trials.

\paragraph{Earnings days carry the signal.} In a post hoc split, every signal's pre-open IC is concentrated on
the \res{earn-n-test} test events with a scheduled earnings report, and there it held in all three periods.
Laya-Fin's IC with $z_{cc}$ on earnings days was \res{lf-earn-vala}, \res{lf-earn-valb} and \res{lf-earn-test} on
val-A, val-B and the test, against \res{lf-rest-test} on the other test events; d1-ultra's was
\res{d1-earn-test} on the test ($t = \res{d1-earn-test-t}$). The fall from validation to test of
Section~\ref{sec:validation} is therefore a fall on the other events. Even on earnings days the signal is priced
at the open: d1-ultra's earnings-day IC on the test is \res{d1-earn-on-test} with $z_{on}$ and
\res{d1-earn-id-test} with $z_{id}$.

\begin{figure}[tbp]
\centering
\includegraphics{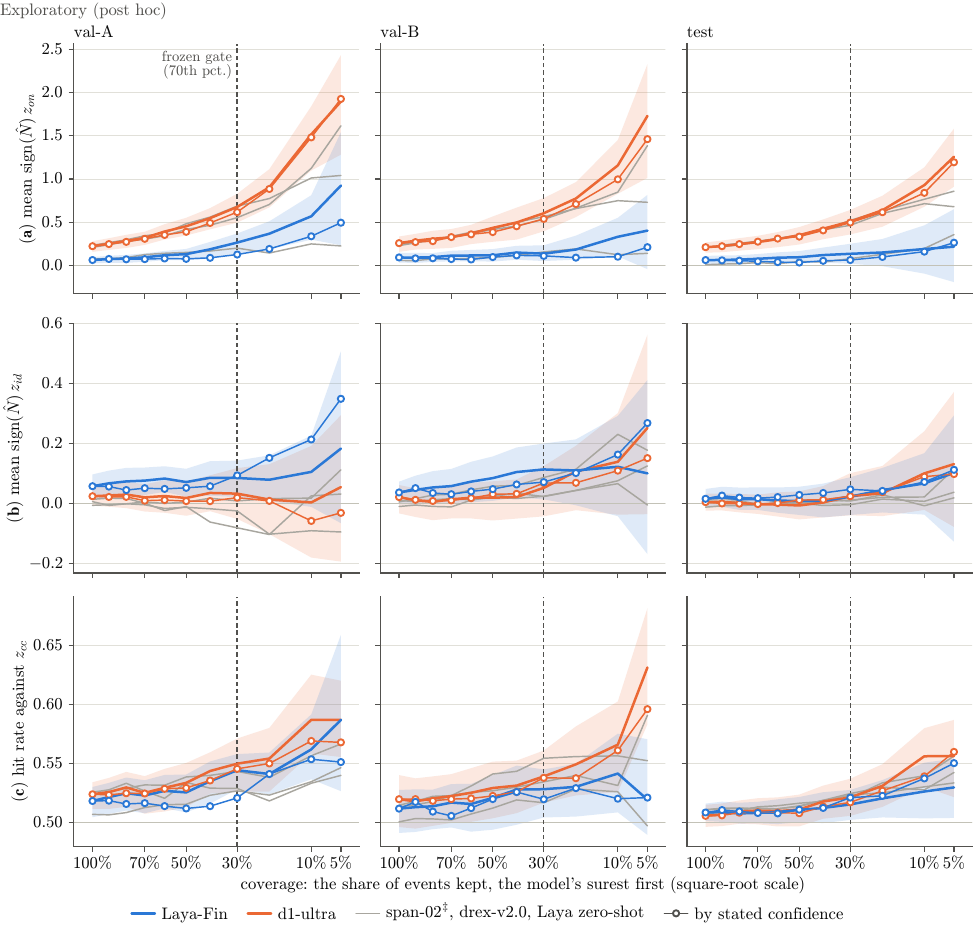}
\caption{\textbf{Exploratory (post hoc): the overnight signal concentrates in d1-ultra's most confident tenth of
pre-open readings in all three periods (mean signed overnight move \res{sel-d1-on-10-vala},
\res{sel-d1-on-10-valb} and \res{sel-d1-on-10-test} typical moves, against \res{sel-d1-on-all-vala},
\res{sel-d1-on-all-valb} and \res{sel-d1-on-all-test} for all events); the move after the open shows no such
concentration.}
Coverage curves: events ranked by conviction $|\Nhat|$ (lines) or by the stated confidence (circles), keeping the
most confident share, with thresholds from val-A; 95\% intervals from the stationary bootstrap over sessions.
(a) Mean $\mathrm{sign}(\Nhat)\cdot z_{on}$; (b) mean $\mathrm{sign}(\Nhat)\cdot z_{id}$; (c) the hit rate
against $z_{cc}$. The dashed line marks the frozen confidence gate (the val-A 70th percentile).
$^{\ddagger}$See Appendix~\ref{app:prereg}.}\label{fig:selective}
\end{figure}

\paragraph{Selective prediction: the overnight signal concentrates in the most confident calls.} We ranked events
post hoc by conviction, the size of $|\Nhat|$ against its val-A distribution, and kept only the top share of them,
which traces a coverage curve \citep{geifman2017} (Figure~\ref{fig:selective}). d1-ultra's mean overnight move in
the direction of its reading, $\mathrm{sign}(\Nhat)\cdot z_{on}$, is \res{sel-d1-on-10-test} typical moves in its
top 10\% on the test, against \res{sel-d1-on-all-test} for all events, and the pattern held in both validation
quarters. After the open, selectivity gains little: no test interval of the top share's mean move after the open
excludes zero at any coverage.

\paragraph{Watching the open adds nothing.} In a post hoc comparison, the reading at 10:00, after the first
half-hour bar, never predicts the rest of the day better than the pre-open reading on val-B or the test, and the
revision from the pre-open to the 10:00 reading has a zero or negative IC with the move from 10:00 to the close
(Laya-Fin \res{watch-lf-rev-test} on the test, $t = \res{watch-lf-rev-test-t}$). The models seem to over-update
toward the first bar, but the comparison involves many statistics, so this exploratory pattern is weak evidence.

\paragraph{Validation-to-test decay, the strongest instance.} A post hoc conditioning study asked whether the
after-the-open signal $\mathrm{sign}(\Nhat)\cdot z_{id}$ is stronger under some condition (the type and timing of
the news, the day, the stock, the models' internals, agreement between models and a cost hurdle) for
\res{cond-pairs} (model, condition) pairs. A strict rule, fixed before its test run, selected the pairs positive
with $t \ge 1$ in both validation quarters and a pooled $t \ge 2$, and \res{cond-selected} passed. On the test,
\res{cond-reversed} of them changed sign and the other \res{cond-positive} are small (Appendix
Figure~\ref{fig:a-conditioning}). Their tradable versions had in-sample gross Sharpe ratios up to
\res{cb-valb-gross-max} on val-B; on the test, none reaches the 90th percentile of its placebo, and every one loses
net of 10 basis points.

\section{Limitations and the forward test}\label{sec:limits}

\paragraph{Limitations.} The sample is the S\&P 500, the largest and most closely followed US stocks; smaller
stocks, where news may diffuse more slowly \citep{hong2000}, may behave differently. The horizon is daily, with one
intraday decision point at 10:00 built from 30-minute bars whose first bar starts at the first trade, not at the
opening auction. The analysis scores the pre-open readings against the whole overnight move and does not split it
at the 09:15 reading, so it cannot say how much of that move happened before the reading. The test period is seven
months long (\res{test-d0-sessions} sessions with D0 events), so small ICs are hard to detect, and the selection
rested on two validation quarters of \res{valb-sessions} sessions each. The trained instruments were trained once,
with one seed and one epoch; the anchor learned the size of moves but not their direction and is uninformative, so
the study says nothing about what fine-tuning adds. The commercial models failed the pre-registered screens and
were released after the sample, so contamination cannot be ruled out. Five of the authors wrote the earlier study
\citep{kargarzadeh2026}, whose 2023--2026 sample overlaps our test period; none of its statistics is a test in
this paper, but its authors had seen aggregate news reactions that include early 2026 before this design was
fixed. The vendor-label baselines rest on labels
produced after the fact. Masking leaves small residues of names and dates, below half a percent of states. The
strategies charge flat costs per side and model no borrow costs or market impact, and the hedge removes market risk
only. Some pre-registered items were not run or not completed: three robustness variants that needed new model
calls (a summary-text variant, vendor tags in the text and a 09:30 news cutoff; the 08:00 cutoff is approximated),
the second and third training seeds, and \res{ablation-failed} rate-limited requests of the ablations, which were
not retried.

\paragraph{The forward test.} The cleanest test of memorization uses news published after every model's
release. Seven hypotheses for such a test were pre-registered on 2026-10-29, before any news from the forward
window was exported and before any forward-period model call; the window runs from the first session after
2026-10-21 for at least three months, including the January--February 2027 earnings season, with the frozen
pipeline run unchanged. They are:
\begin{itemize}[leftmargin=*, itemsep=1pt]
\item \textbf{F1, memorization:} for each commercial model, the change in IC from the test to the forward window
  minus the same change for the clean baselines is negative (Holm across the three models);
\item \textbf{F2, entity against event knowledge:} company identification on masked forward states against
  masked test states, and against the matched baseline;
\item \textbf{F3, the conviction book:} the pre-open book restricted to each model's top tenth of $|\Nhat|$
  (thresholds fixed from val-A) has a positive net Sharpe ratio;
\item \textbf{F4, priced at the open:} the frozen H1 and H1b on the forward window;
\item \textbf{F5, over-updating:} the IC of the revision after the first half hour with the rest of the day's
  move (two-sided);
\item \textbf{F6, overconfidence:} the commercial models' expected calibration error exceeds Laya-Fin's (Holm
  across the three models);
\item \textbf{F7, conviction concentration:} the mean signed overnight move of the top tenth by conviction
  exceeds that of all events (Holm across four models).
\end{itemize}
F5 and F7 turn exploratory results of Section~\ref{sec:exploratory}, and F6 a descriptive comparison of
Section~\ref{sec:calibration}, into confirmatory tests. The forward test has not been run. The robustness runs of
the anchor (seeds 17 and 23 and a
longer training run), fixed before unblinding, have not been run either; they are robustness only and cannot
change any verdict here.

\section{Conclusion}\label{sec:conclusion}

AI decision models read financial news well enough to line up with how a stock opens, and not well enough to
forecast what it does after that. A fine-tuned open-weights model that looked convincing on two validation
quarters did not survive a blinded test. The commercial models kept a signal, but only in the overnight gap,
where the market had already priced the news before anyone could trade on it at the open, as an efficient market
would \citep{fama1970}. The models could not tell an overreaction from an underreaction, their calls of
underreaction never caught up, and no pre-registered strategy survived costs and selection. Out of sample, news
explains a fraction of a percent of a stock's own daily variance, and the commercial models' stated
probabilities overstate their accuracy.

Three recommendations follow for anyone who backtests language-model readings of news. First, split the return at
the open: a pre-open signal can be traded only if it predicts the move after the open, and a close-to-close
backtest credits it with a move nobody could trade. Second, measure recognition, not just masking: report how
often the model names the masked company against a matched baseline, and where possible test on news published
after the model's release, as our pre-registered forward test will. Third, freeze and count, as OLIVIA prescribes
(Table~\ref{tab:olivia}): select
on validation data only, freeze the plan by hash, and log every trial and deflate for them; the strongest
exploratory pattern we
found, with in-sample Sharpe ratios up to \res{cb-valb-gross-max}, reversed on the test. The market reads the news
first.

\section*{Acknowledgments and statements}
\paragraph{Acknowledgments.} The NewsWitch news corpus was provided by Zanista AI Ltd, and the cleaned dataset was
prepared at TailState Intelligence Ltd.

\paragraph{Funding.} This research received no external funding.

\paragraph{Competing interests.} Arman Khaledian and Navid Parvini are affiliated with Zanista AI Ltd, which
develops NewsWitch, the commercial news feed used in this study, whose sentiment and topic labels are
evaluated here as baselines. Alireza Kargarzadeh is affiliated with TailState Intelligence Ltd, which
prepared the dataset. The other authors declare no competing interests. The views expressed are the
authors' own and do not necessarily reflect those of their employers.

\paragraph{Use of AI tools.} An AI coding and research assistant (Claude, by Anthropic, used through Claude Code)
implemented and ran the pre-registered pipeline under the authors' direction and drafted this manuscript from the
verified outputs. The authors take full responsibility for the content.

\renewcommand{\bibfont}{\small}
\setlength{\bibsep}{2pt plus 1pt}
\bibliographystyle{plainnat}
\bibliography{references}

\clearpage
\appendix
\renewcommand{\thefigure}{A\arabic{figure}}\setcounter{figure}{0}
\renewcommand{\thetable}{A\arabic{table}}\setcounter{table}{0}
\renewcommand{\theequation}{A\arabic{equation}}\setcounter{equation}{0}

\section{Pre-registration and deviations}\label{app:prereg}

The pre-registration fixes every definition, constant, split, hypothesis and selection rule. After it was
written, changes could only be appended to its deviations log, each with its date, the section it touches, the
reason and whether it came before or after the freeze; the freeze record holds the log's length at the freeze, and
the lock refused to release the test outcomes unless the log was an unchanged prefix of the current one. The full
log is in the replication package. Most entries make precise what the pre-registration left open (how a filter
reaches its budget, which events a sample holds, how a variable is scaled) and were written before the freeze.
The material deviations are these.
\begin{itemize}[leftmargin=*, itemsep=1pt]
\item \emph{Amendments before any data.} Convergence to the news-implied value, a wider question battery and the
  firm-level explanation analysis were added on the day the pre-registration was written.
\item \emph{Masking.} The masking rules were widened several times before the freeze, after independent checks
  found residual names, dates and price moves; non-English articles, move explainers and running market stories
  were excluded; residues below half a percent of states were accepted and disclosed.
\item \emph{The commercial models} failed the pre-registered screens. Instead of dropping them, the three with
  the highest second-stage IC were kept as screen-failing, model-specific comparators (before the freeze).
\item \emph{Training.} The trained instruments are one seed and one epoch each, not the pre-registered three-seed
  ensemble, because of the compute available; the other seeds are robustness runs, fixed before unblinding and
  not yet run.
\item \emph{H1b} (the IC with the move after the open) was added to the primary family before the freeze, because
  the strategy that enters at the open needs it.
\item \emph{Company identification} was re-run with activity-matched distractors before the freeze, because a
  guess from news activity alone beat chance in the original design.
\item \emph{Items not run.} Three robustness variants needed new model calls and were not run (a summary-text
  variant, vendor tags in the text and a 09:30 news cutoff); the 08:00 cutoff is approximated from time stamps.
  Of the ablation requests, \res{ablation-failed} were rate-limited and not retried.
\item \emph{The canary.} After the freeze, a stability check that repeats earlier requests fired for span-02
  before production. Its build string was unchanged and the drift was close to its own repeat noise, so production
  went ahead under a rule recorded before the end-of-production check: a model counts as updated only if its build
  changes or two independent checks fire. At the end of production neither condition held for any model. The
  figures and tables mark span-02 with $^{\ddagger}$ for this reason.
\item \emph{The price-only baseline's post-close predictions} were moved after unblinding, because one of their
  inputs, the day's overnight move, is a test-period outcome; the model was unchanged and first reproduced its
  validation predictions exactly.
\item \emph{Analysis and strategies.} The operationalization of the frozen specification and of the strategies
  was written before the outcomes were read; three code fixes found when the code first met the test data changed
  no pre-registered number.
\item \emph{Exploratory work after unblinding} is dated in the log. The conditioning study's validation selection
  was committed before its single test run; its two definitional choices (the VIX close of the previous session,
  read from a registered copy of the market data, and payroll release dates added from the Bureau of Labor
  Statistics) were made on validation data.
\item \emph{The forward test} was pre-registered after the paper's analyses, before any forward-period news was
  exported and before any forward-period model call (Section~\ref{sec:limits}).
\end{itemize}

\section{Notation}\label{app:notation}

{\small
\begin{longtable}{@{}>{\raggedright\arraybackslash}p{1.35in}>{\raggedright\arraybackslash}p{4.95in}@{}}
\caption{Notation.}\label{tab:notation} \\
\toprule
Symbol & Meaning \\
\midrule
\endfirsthead
\multicolumn{2}{@{}l}{\textit{Table \ref{tab:notation}, continued}} \\
\toprule
Symbol & Meaning \\
\midrule
\endhead
\bottomrule
\endlastfoot
D0, D1, D2 & Decision points: 09:15~ET before the open (pre-open reading); 10:00~ET after the first 30-minute bar
and after the close (post-reaction readings). \\
val-A, val-B, test & 2025-07-01 to 2025-09-30; 2025-10-01 to 2025-12-31; 2026-01-02 to 2026-07-31. \\
$r^{cc}, r^{on}, r^{id}$ & Close-to-close, overnight (close to open) and intraday (open to close) returns, with
$1 + r^{cc} = (1 + r^{on})(1 + r^{id})$; Equation~\eqref{eq:split}. \\
$\AR^{k}$, $\sigma^{k}_{i,t-1}$ & Abnormal return (market and leave-one-out sector removed, betas over 252
sessions to $t-1$); its ex-ante scale (1.4826 times the median absolute deviation over 126 sessions to $t-1$,
floored at the day's cross-sectional 10th percentile). \\
$z^{k}$; $z_{cc}, z_{on}, z_{id}$ & $\AR^{k}/\sigma^{k}_{i,t-1}$, the move in typical daily moves (clipped at
$\pm 10$). \\
$y_0, y_1, y_2$ & Outcomes after D0, D1 and D2: $z_{id}$; the move from 10:00 to the close; the move from the next
open to the close of $t+5$, each over its ex-ante volatility. \\
$e_5$ & The move from the previous close to the close of $t+4$ over $\sigma^{cc}\sqrt{5}$: the training label and
the calibration target. \\
$G$, $\Glvl$ & The observed reaction at D1 or D2 in typical-move units; its level version, divided by
$\sigma^{cc}$ without a square-root-of-time factor. \\
$\Nhat$, $\Ntilde$ & The news-implied move $\sum_k p_k m_k$ over the nine levels of the pre-open reading; its
calibrated version (an isotonic map to $e_5$ fitted on val-A), which ranks the pre-open strategy. \\
$V$, $U$ & The news-implied value (an isotonic map of $\Nhat$ to the level move to $t+4$, fitted on val-A); the
remaining gap $V - \Glvl$ at D1 or D2. \\
$J$ & The post-reaction judgment $P(\text{much too small}) + P(\text{too small}) - P(\text{too large}) -
P(\text{much too large})$. \\
Under, Fair, Over & Post-reaction classes from $(\Nhat - G)\,\mathrm{sign}(\Nhat)$ for $|\Nhat| \ge 0.5$: Under
above 0.5, Over below $-0.5$, Fair between. \\
$\CARlvl_{h}$ & The level cumulative abnormal return from the previous close to rung $h$. \\
$\kappa_h$, $\theta_V$ & The convergence slope at rung $h$ (Equation~\eqref{eq:ladder}); the reliability of $V$
(Equation~\eqref{eq:attenuation}). \\
$\lambda_h$, $\psi_h$, $h^{*}$ & The share of the gap closed by rung $h$, the convergence to a shrunken value and
the half-life (Equation~\eqref{eq:gap}). \\
$\IC_t$, $\overline{\IC}$ & The daily Spearman information coefficient and its mean over sessions
(Equation~\eqref{eq:ic}). \\
$R^{2}_{\mathrm{OOS}}$, $\Delta R^{2}$ & The out-of-sample $R^2$ and a step's increment
(Equation~\eqref{eq:r2}). \\
$\mathrm{Acc}_m$, $p_0$ & Company-identification accuracy and the matched baseline
(Equation~\eqref{eq:memorization}). \\
ECE & Expected calibration error of the top class of three (a fall, no move or a rise, split at $\pm 0.5$
typical moves against $e_5$), 15 equal-width bins. \\
DSR, $\mathrm{SR}_0$ & The deflated-Sharpe probability and its benchmark (Equation~\eqref{eq:dsr}). \\
H1, H1b, H2, H3, H4 & Primary hypotheses (Holm at 0.05). \\
H5, H6, H7, H2-C & Secondary hypotheses (Benjamini--Hochberg at $q = 0.10$): explaining moves, style, the
instruments, convergence. \\
F1 to F7 & The forward test's hypotheses (Section~\ref{sec:limits}). \\
\end{longtable}
}

\section{The deflated Sharpe ratio}\label{app:dsr}

For a book with $T$ daily net returns, a per-period Sharpe ratio $\widehat{\mathrm{SR}}$, skewness
$\hat\gamma_3$ and kurtosis $\hat\gamma_4$, the deflated-Sharpe probability \citep{bailey2014} is computed as
\begin{equation}\label{eq:dsr}
\mathrm{DSR} = \Phi\!\left(\frac{(\widehat{\mathrm{SR}} - \mathrm{SR}_0)\sqrt{T-1}}
{\sqrt{1 - \hat\gamma_3\,\widehat{\mathrm{SR}} + \tfrac{\hat\gamma_4 - 1}{4}\,\widehat{\mathrm{SR}}^{2}}}\right),
\qquad
\mathrm{SR}_0 = \sqrt{\mathcal{V}}\left[(1 - \gamma)\,\Phi^{-1}\!\left(1 - \frac{1}{N}\right)
+ \gamma\,\Phi^{-1}\!\left(1 - \frac{1}{N\mathrm{e}}\right)\right],
\end{equation}
where $\Phi$ is the standard normal distribution function, $\gamma$ the Euler--Mascheroni constant, $N$ the
number of trials and $\mathcal{V} = (1 + \widehat{\mathrm{SR}}^{2}/2)/(T-1)$ the variance of a Sharpe-ratio
estimate \citep{lo2002}; the analytic variance is used because most of the ledger's other trials are information
coefficients, recalls and losses, not Sharpe ratios. $\mathrm{SR}_0$ is the Sharpe ratio that the best of $N$
independent trials without skill would reach in expectation, so it rises with $N$: a book must beat it, not zero.
With the \res{ntrials-strat} trials of the ledger when the pre-registered strategies were evaluated and
\res{sconv-ndays} test days, $\mathrm{SR}_0$ is \res{dsr-sr0} per day for the convergence strategy, far above any
book's daily Sharpe ratio. The exploratory books of Section~\ref{sec:exploratory} were deflated with all
\res{ntrials-total} trials of the ledger after the exploratory study, which raises the bar further.

\section{Additional pre-registered results}\label{app:results}

\begin{figure}[htbp]
\centering
\includegraphics{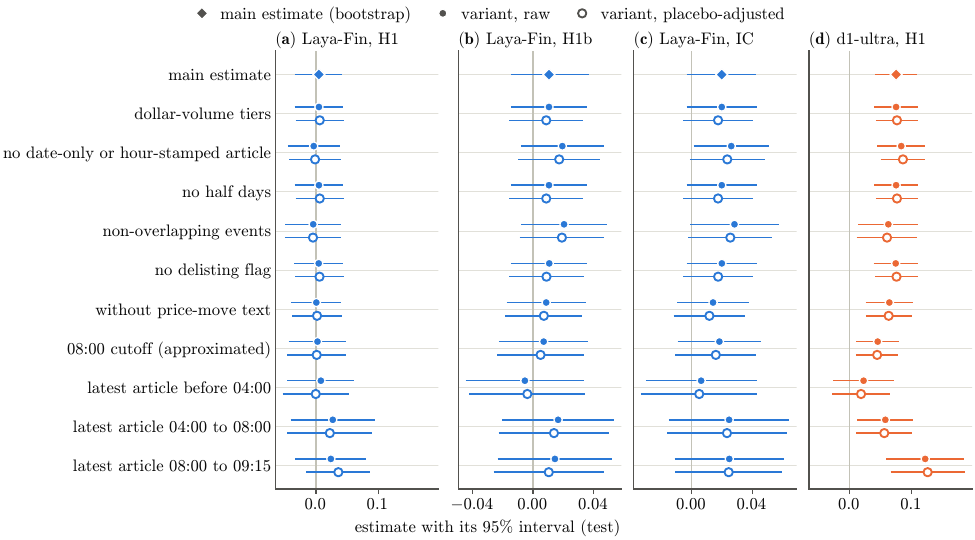}
\caption{\textbf{No pre-registered robustness variant changes a conclusion: Laya-Fin's H1 and H1b stay within
their intervals of zero, and d1-ultra's H1 stays positive in every variant, from \res{d1-h1-pre04} when the
latest article precedes 04:00 to \res{d1-h1-0815} when it arrives between 08:00 and 09:15.} Test period, D0
events. Each variant raw (filled) and with placebo-adjusted targets (hollow), below the main estimate (diamond,
from the stationary block bootstrap), with 95\% intervals ($\pm 1.96$ Newey--West standard errors for the
variants). (a) and (d), both H1, share a scale; (b) and (c) have their own.}\label{fig:a-robustness}
\end{figure}

This appendix gives the pre-registered results that the main text cites but does not show: every model's IC
and cost (Table~\ref{tab:a-model-comparison}), the robustness variants (Figure~\ref{fig:a-robustness}), the
heterogeneity cuts (Table~\ref{tab:a-heterogeneity}), the secondary hypotheses (Table~\ref{tab:a-secondary}), the
convergence ladder with what it identifies and the gap-closure test (Appendix~\ref{app:convergence} and
Table~\ref{tab:convergence}), and calibration (Table~\ref{tab:a-calibration} and Figure~\ref{fig:a-reliability}).
Further tables and figures are in the replication package.

\begin{table}[htbp]
\centering
\footnotesize
\begin{threeparttable}
\caption{Model comparison: every val-B IC with $z_{cc}$ that is clearly positive falls on the test (Laya-Fin 0.065 to 0.020); the cost and the latency.}\label{tab:a-model-comparison}
\begin{tabular}{@{}>{\raggedright\arraybackslash}p{1.75in}S[table-format=-1.3]S[table-format=-1.2]S[table-format=-1.3]S[table-format=-1.2]S[table-format=5.0]S[table-format=1.3]S[table-format=4.0]@{}}
\toprule
Model & {IC val-B} & {$t$} & {IC test} & {$t$} & {Events} & {USD/1,000} & {p50 (ms)} \\
\midrule
Laya-Fin & 0.065 & 3.64 & 0.020 & 1.67 & 13748 & 0.000 & {---} \\
anchor & -0.020 & -1.52 & 0.008 & 0.83 & 13748 & 0.000 & {---} \\
Laya zero-shot & 0.029 & 1.94 & 0.005 & 0.37 & 13748 & 0.000 & 536 \\
d1-ultra & 0.097 & 5.38 & 0.039 & 3.23 & 13748 & 0.063 & 288 \\
span-02$^{\ddagger}$ & 0.099 & 5.83 & 0.041 & 3.10 & 13748 & 0.028 & 2262 \\
drex-v2.0 & 0.075 & 4.10 & 0.044 & 3.70 & 13748 & 0.045 & 680 \\
vendor sentiment$^{\dagger}$ & 0.087 & 4.39 & 0.047 & 3.36 & 13748 & {---} & {---} \\
bge-ridge & 0.002 & 0.10 & 0.004 & 0.45 & 13748 & {---} & {---} \\
tags-GBM$^{\dagger}$ & 0.108 & 6.03 & 0.055 & 3.91 & 13748 & {---} & {---} \\
LLM baseline & 0.052 & 1.93 & 0.001 & 0.02 & 1977 & {---} & {---} \\
price-only & 0.004 & 0.33 & -0.003 & -0.26 & 13748 & {---} & {---} \\
Laya-Fin, LLM's 2,000-event subsample & {---} & {---} & -0.012 & -0.37 & 1977 & {---} & {---} \\
LLM baseline, its subsample & {---} & {---} & 0.001 & 0.02 & 1977 & {---} & {---} \\
price-only (D2, IC with $y_2$) & {---} & {---} & 0.012 & 1.18 & 17188 & {---} & {---} \\
\bottomrule
\end{tabular}
\begin{tablenotes}[flushleft]\footnotesize
\item ---: the source table has no value for the row; the statistic does not apply to it or was not computed.
\item $^{\dagger}$Vendor-label baseline (vendor sentiment, tags-GBM): the labels were produced retrospectively, so its IC is a possibly look-ahead-contaminated upper bound, not the benchmark; the clean benchmarks are bge-ridge, price-only, the LLM baseline and the random-signal placebo.
\item $^{\ddagger}$span-02: after the freeze, a stability check that repeats earlier requests (the canary) fired for it; its build was unchanged, and production went ahead under a rule recorded in the deviations log.
\item The anchor's readings are uninformative (near-constant outputs after one epoch); its rows are reported, not read as nulls.
\item The API models fail the pre-registered contamination screens, so their results are model-specific.
\end{tablenotes}
\end{threeparttable}
\end{table}

\begin{table}[htbp]
\centering
\small
\begin{threeparttable}
\caption{Heterogeneity of H2: no cut differs (Laya-Fin's Wald $p$ from 0.16 to 0.90).}\label{tab:a-heterogeneity}
\begin{tabular}{@{}lS[table-format=5.0]S[table-format=1.2]S[table-format=1.3]S[table-format=1.3]S[table-format=-1.3]S[table-format=1.3, table-space-text-pre=(, table-space-text-post=)]S[table-format=-1.3]S[table-format=1.3, table-space-text-pre=(, table-space-text-post=)]@{}}
\toprule
Model, cut & {Events} & {Wald $F$} & {$p$} & {BH $q$} & {$\Delta b$} & {SE} & {$\Delta c$} & {SE} \\
\midrule
Laya-Fin, Top 100 & 4516 & 1.83 & 0.164 & 0.741 & 0.490 & {(}0.281{)} & -0.025 & {(}0.021{)} \\
Laya-Fin, 101 to 250 & 5121 & 0.10 & 0.904 & 0.904 & -0.008 & {(}0.258{)} & -0.007 & {(}0.016{)} \\
Laya-Fin, 251 to 500 & 7551 & 1.71 & 0.185 & 0.741 & -0.326 & {(}0.253{)} & 0.023 & {(}0.016{)} \\
Laya-Fin, quantified & 15030 & 0.66 & 0.521 & 0.833 & 0.470 & {(}0.609{)} & 0.022 & {(}0.026{)} \\
Laya-Fin, small move & 9795 & 0.29 & 0.751 & 0.859 & -0.193 & {(}0.255{)} & 0.005 & {(}0.035{)} \\
Laya-Fin, press wire & 9655 & 0.75 & 0.477 & 0.833 & -0.058 & {(}0.330{)} & -0.024 & {(}0.020{)} \\
Laya-Fin, low usual coverage & 8387 & 0.44 & 0.645 & 0.859 & 0.040 & {(}0.247{)} & 0.014 & {(}0.016{)} \\
Laya-Fin, busy day & 7805 & 0.93 & 0.398 & 0.833 & -0.194 & {(}0.263{)} & -0.019 & {(}0.019{)} \\
anchor, Top 100 & 4516 & 2.18 & 0.117 & 0.469 & 6.973 & {(}4.120{)} & -0.023 & {(}0.020{)} \\
anchor, 101 to 250 & 5121 & 0.52 & 0.594 & 0.679 & -2.559 & {(}2.719{)} & -0.007 & {(}0.016{)} \\
anchor, 251 to 500 & 7551 & 1.18 & 0.312 & 0.679 & -2.500 & {(}3.035{)} & 0.021 & {(}0.015{)} \\
anchor, quantified & 15030 & 2.84 & 0.062 & 0.469 & 7.109 & {(}3.308{)} & 0.024 & {(}0.026{)} \\
anchor, small move & 9795 & 0.01 & 0.992 & 0.992 & 0.018 & {(}2.650{)} & 0.004 & {(}0.035{)} \\
anchor, press wire & 9655 & 0.79 & 0.457 & 0.679 & -1.650 & {(}3.084{)} & -0.024 & {(}0.020{)} \\
anchor, low usual coverage & 8387 & 0.75 & 0.477 & 0.679 & 2.272 & {(}2.652{)} & 0.014 & {(}0.016{)} \\
anchor, busy day & 7805 & 0.58 & 0.561 & 0.679 & 1.140 & {(}2.570{)} & -0.020 & {(}0.019{)} \\
d1-ultra, Top 100 & 4516 & 1.10 & 0.337 & 0.719 & 0.059 & {(}0.060{)} & -0.024 & {(}0.020{)} \\
d1-ultra, 101 to 250 & 5121 & 0.88 & 0.418 & 0.719 & 0.063 & {(}0.055{)} & -0.010 & {(}0.016{)} \\
d1-ultra, 251 to 500 & 7551 & 3.36 & 0.038 & 0.301 & -0.095 & {(}0.050{)} & 0.026 & {(}0.015{)} \\
d1-ultra, quantified & 15030 & 0.53 & 0.591 & 0.719 & 0.014 & {(}0.062{)} & 0.023 & {(}0.027{)} \\
d1-ultra, small move & 9795 & 0.15 & 0.857 & 0.857 & 0.023 & {(}0.044{)} & 0.002 & {(}0.035{)} \\
d1-ultra, press wire & 9655 & 0.67 & 0.515 & 0.719 & -0.021 & {(}0.049{)} & -0.022 & {(}0.020{)} \\
d1-ultra, low usual coverage & 8387 & 2.52 & 0.084 & 0.336 & 0.096 & {(}0.046{)} & 0.010 & {(}0.016{)} \\
d1-ultra, busy day & 7805 & 0.47 & 0.629 & 0.719 & 0.002 & {(}0.043{)} & -0.019 & {(}0.019{)} \\
\bottomrule
\end{tabular}
\begin{tablenotes}[flushleft]\footnotesize
\item Wald test of $\hat N\cdot 1[\mathrm{cut}]$ and $G\cdot 1[\mathrm{cut}]$ added to H2 at D2; BH over each model's eight cuts. Events: the D2 events in the cut.
\item The anchor's readings are uninformative (near-constant outputs after one epoch); its rows are reported, not read as nulls.
\item The API models fail the pre-registered contamination screens, so their results are model-specific.
\end{tablenotes}
\end{threeparttable}
\end{table}

\begin{table}[htbp]
\centering
\small
\begin{threeparttable}
\caption{Secondary hypotheses: only H5 survives Benjamini--Hochberg for Laya-Fin ($q$ = 0.058).}\label{tab:a-secondary}
\begin{tabular}{@{}lS[table-format=-1.4]cS[table-format=1.4, table-space-text-pre=(, table-space-text-post=)]S[table-format=1.3, table-space-text-pre={$<$}]S[table-format=1.3, table-space-text-pre={$<$}]S[table-format=1.3]@{}}
\toprule
Model, test & {Estimate} & 95\% CI & {SE} & {$p$} & {BH $q$} & {Pooled $q$} \\
\midrule
Laya-Fin, H5 & 0.0025 & --- & {---} & 0.010 & 0.058 & 0.078 \\
Laya-Fin, H6 & {---} & --- & {---} & {---} & {---} & {---} \\
Laya-Fin, H7 & 0.0146 & [$-0.0068$, $0.0363$] & {---} & 0.097 & 0.238 & 0.317 \\
Laya-Fin, H2-C, $\kappa$ at $h^{*}$ & 0.2588 & --- & {(}0.2299{)} & 0.131 & 0.238 & 0.317 \\
Laya-Fin, H2-C, $\kappa$ increase & 0.0182 & [$-0.3614$, $0.4903$] & {---} & 0.467 & 0.561 & 0.748 \\
Laya-Fin, H2-C, $\lambda_{t+5}$ & 0.0265 & --- & {(}0.0264{)} & 0.159 & 0.238 & 0.317 \\
Laya-Fin, H2-C, naive tracking & 1.0007 & [$0.9995$, $1.0018$] & {---} & 0.867 & 0.867 & 0.946 \\
Laya-Fin, A3.4 & {---} & --- & {---} & {---} & {---} & {---} \\
anchor, H5 & 0.0003 & --- & {---} & 0.072 & 0.358 & {---} \\
anchor, H6 & {---} & --- & {---} & {---} & {---} & {---} \\
anchor, H7 & 0.0024 & [$-0.0294$, $0.0355$] & {---} & 0.442 & {---} & {---} \\
anchor, H2-C, $\kappa$ at $h^{*}$ & 0.0798 & --- & {(}0.1574{)} & 0.307 & 0.511 & {---} \\
anchor, H2-C, $\kappa$ increase & -0.0031 & [$-0.2669$, $0.4215$] & {---} & 0.514 & 0.643 & {---} \\
anchor, H2-C, $\lambda_{t+5}$ & 0.0270 & --- & {(}0.0266{)} & 0.156 & 0.390 & {---} \\
anchor, H2-C, naive tracking & 1.0020 & [$1.0002$, $1.0036$] & {---} & 0.992 & 0.992 & {---} \\
anchor, A3.4 & {---} & --- & {---} & {---} & {---} & {---} \\
d1-ultra, H5 & 0.0082 & --- & {---} & {$<$}0.001 & {$<$}0.001 & {---} \\
d1-ultra, H6 & 0.0104 & --- & {(}0.0192{)} & 0.705 & 0.946 & 0.940 \\
d1-ultra, H7 & {---} & --- & {---} & {---} & {---} & {---} \\
d1-ultra, H2-C, $\kappa$ at $h^{*}$ & 0.3645 & --- & {(}0.2152{)} & 0.046 & 0.163 & {---} \\
d1-ultra, H2-C, $\kappa$ increase & -0.1805 & [$-0.5343$, $0.2158$] & {---} & 0.834 & 0.946 & {---} \\
d1-ultra, H2-C, $\lambda_{t+5}$ & 0.0216 & --- & {(}0.0262{)} & 0.206 & 0.480 & {---} \\
d1-ultra, H2-C, naive tracking & 1.0005 & [$0.9993$, $1.0017$] & {---} & 0.781 & 0.946 & {---} \\
d1-ultra, A3.4 & -0.0619 & --- & {(}0.0382{)} & 0.946 & 0.946 & 0.946 \\
\bottomrule
\end{tabular}
\begin{tablenotes}[flushleft]\footnotesize
\item ---: the source table has no value for the row; the statistic does not apply to it or was not computed.
\item BH at $q = 0.10$ within each instrument; the pooled family adds d1-ultra's H6 and A3.4 to Laya-Fin's. The anchor's H7 is reported outside its family, so it has no $q$. Each test has either an SE or a 95\% CI, as the frozen analysis reports it.
\item A3.4: the pre-registered regression of next-week persistence on the explanation answer. Empty rows: H6 needs the tone and materiality answers and A3.4 the explanation answer, which the trained models do not give; H7 compares a trained model with Laya zero-shot.
\item The anchor's readings are uninformative (near-constant outputs after one epoch); its rows are reported, not read as nulls.
\item The API models fail the pre-registered contamination screens, so their results are model-specific.
\end{tablenotes}
\end{threeparttable}
\end{table}

\subsection{The convergence ladder: what it identifies, and gap closure}\label{app:convergence}

\paragraph{What the ladder identifies.} A slope below one in Equation~\eqref{eq:ladder} does not by itself mean
that the market underreacts, because $V$ is a model's estimate. A standard errors-in-variables argument
\citep{hausman2001} makes this precise; we add it as an interpretation, not as a new test. Suppose the model's
value is a noisy measure of the true news-implied value, $V = V^{*} + e$, with classical error: $e$ has mean zero
and is independent of $V^{*}$ and of the regression error, after the fixed effects are partialled out. Then the
slope converges to
\begin{equation}\label{eq:attenuation}
\hat\kappa_h \;\to\; \theta_V\,\kappa^{*}_h,
\qquad
\theta_V = \frac{\operatorname{Var}(V^{*})}{\operatorname{Var}(V^{*}) + \operatorname{Var}(e)} \le 1,
\end{equation}
where $\kappa^{*}_h$ is the slope on the true value. Two consequences follow. First, the \emph{level} of
$\kappa_h$ mixes market efficiency with model noise, so $\kappa_h < 1$ alone is not evidence of underreaction.
Second, the \emph{shape} of the ladder is identified up to the same factor: on a common set of events,
$\hat\kappa_{h'} - \hat\kappa_h \to \theta_V(\kappa^{*}_{h'} - \kappa^{*}_h)$. The factor is positive, so the sign
is preserved: a rise after the decision point means drift toward the true value and a fall means reversal. This
is why the pre-registered tests read the ladder's shape (the increase $\kappa_{t+20} - \kappa_{O_t}$ and the gap
closure below) rather than its level. Table~\ref{tab:convergence} gives every rung for every model.

\paragraph{Gap closure.} The gap between the model's value and the price after the reaction,
$U_{i,\tau} = V_i - \Glvl_{i,\tau}$ at decision point $\tau$ (D1 or D2), should close if the price converges to
the value; $\Glvl$ is the observed reaction divided by $\sigma^{cc}$ without a square-root-of-time factor, so that
moves at different horizons add up. With $\AR(\tau \to h)$ the abnormal move from the decision point to rung $h$,
\begin{equation}\label{eq:gap}
\frac{\AR_{i}(\tau \to h)}{\sigma^{cc}_{i}} = \alpha_{d(t)\times\tau(i)} + \lambda_h\,U_{i,\tau} + \psi_h V_i
+ \varepsilon_{i,h},
\qquad
h^{*} = \min\bigl\{h : \lambda_h \ge \tfrac{1}{2}\bigr\},
\end{equation}
where $\lambda_h$ is the share of the gap closed by rung $h$, $\psi_h$ captures convergence to a shrunken value,
and the half-life $h^{*}$ is ``not reached'' when no rung qualifies. The gap does not close. After the close,
Laya-Fin's $\lambda_h$ is \res{lf-lambda-t1} at $t+1$, \res{lf-lambda-t5} at $t+5$ ($p = \res{lf-lambda-t5-p}$)
and \res{lf-lambda-t20} at $t+20$; from 10:00 every $\lambda_h$ is at or below zero; and the half-life is not
reached within $t+20$ for any model. The $\lambda_h$ are almost the same for every model because $V$ is small
next to the day's move, so that $U \approx -\Glvl$: $\lambda_h$ measures how much of the day's move reverses, not
convergence to $V$. The tracking error between the price and the model's value is no smaller than that of a
naive value of zero (ratio \res{lf-tracking-t20} at $t+20$, $p = \res{lf-tracking-t20-p}$). None of the
convergence tests is supported after Benjamini--Hochberg (Table~\ref{tab:a-secondary}).

\begin{table}[htbp]
\centering
\small
\setlength{\tabcolsep}{4pt}
\begin{threeparttable}
\caption{Convergence to the news-implied value $V$: prices reach part of $V$ on day $t$ ($\kappa$ at the close 0.36 for Laya-Fin and 0.59 for d1-ultra), then partly fade; the gap's half-life is not reached within $t{+}20$.}\label{tab:convergence}
\begin{tabular}{@{}lS[table-format=1.2, table-space-text-pre=(, table-space-text-post=)]S[table-format=-1.2, table-space-text-pre=(, table-space-text-post=)]S[table-format=-1.3, table-space-text-pre=(, table-space-text-post=)]S[table-format=-1.3, table-space-text-pre=(, table-space-text-post=)]S[table-format=-1.2, table-space-text-pre=(, table-space-text-post=)]S[table-format=-1.3, table-space-text-pre=(, table-space-text-post=)]S[table-format=-1.2, table-space-text-pre=(, table-space-text-post=)]S[table-format=-1.3, table-space-text-pre=(, table-space-text-post=)]S[table-format=1.0]@{}}
\toprule
 & {$O_t$} & {10:00} & {$C_t$} & {$t{+}1$} & {$t{+}2$} & {$t{+}5$} & {$t{+}10$} & {$t{+}20$} & {Half-life} \\
\midrule
\multicolumn{10}{@{}l}{\textit{Panel A: convergence ladder $\kappa_h$ (D0 events)}} \\
Laya-Fin & 0.24 & 0.27 & 0.36 & 0.36 & 0.38 & 0.26 & 0.21 & 0.26 & {---} \\
 & {(}0.07{)} & {(}0.08{)} & {(}0.09{)} & {(}0.10{)} & {(}0.11{)} & {(}0.14{)} & {(}0.18{)} & {(}0.23{)} & {---} \\
d1-ultra & 0.55 & 0.61 & 0.59 & 0.54 & 0.47 & 0.36 & 0.25 & 0.36 & {---} \\
 & {(}0.08{)} & {(}0.09{)} & {(}0.08{)} & {(}0.09{)} & {(}0.11{)} & {(}0.13{)} & {(}0.15{)} & {(}0.22{)} & {---} \\
anchor & 0.08 & -0.02 & -0.04 & -0.04 & -0.03 & -0.06 & -0.03 & 0.08 & {---} \\
 & {(}0.08{)} & {(}0.08{)} & {(}0.07{)} & {(}0.09{)} & {(}0.09{)} & {(}0.10{)} & {(}0.11{)} & {(}0.16{)} & {---} \\
\addlinespace
\multicolumn{10}{@{}l}{\textit{Panel B: gap closure $\lambda_h$ and the half-life}} \\
Laya-Fin, D2 & {---} & {---} & {---} & -0.007 & {---} & 0.026 & {---} & 0.050 & {$\infty$} \\
 & {---} & {---} & {---} & {(}0.012{)} & {---} & {(}0.026{)} & {---} & {(}0.051{)} & {---} \\
Laya-Fin, D1 & {---} & {---} & -0.011 & -0.024 & {---} & -0.002 & {---} & -0.013 & {$\infty$} \\
 & {---} & {---} & {(}0.013{)} & {(}0.020{)} & {---} & {(}0.034{)} & {---} & {(}0.062{)} & {---} \\
d1-ultra, D2 & {---} & {---} & {---} & -0.009 & {---} & 0.022 & {---} & 0.049 & {$\infty$} \\
 & {---} & {---} & {---} & {(}0.012{)} & {---} & {(}0.026{)} & {---} & {(}0.051{)} & {---} \\
d1-ultra, D1 & {---} & {---} & -0.013 & -0.028 & {---} & -0.012 & {---} & -0.020 & {$\infty$} \\
 & {---} & {---} & {(}0.013{)} & {(}0.020{)} & {---} & {(}0.034{)} & {---} & {(}0.062{)} & {---} \\
anchor, D2 & {---} & {---} & {---} & -0.007 & {---} & 0.027 & {---} & 0.052 & {$\infty$} \\
 & {---} & {---} & {---} & {(}0.012{)} & {---} & {(}0.027{)} & {---} & {(}0.051{)} & {---} \\
anchor, D1 & {---} & {---} & -0.012 & -0.025 & {---} & -0.003 & {---} & -0.013 & {$\infty$} \\
 & {---} & {---} & {(}0.013{)} & {(}0.020{)} & {---} & {(}0.034{)} & {---} & {(}0.062{)} & {---} \\
\bottomrule
\end{tabular}
\begin{tablenotes}[flushleft]\footnotesize
\item Panel A: $\kappa_h$ is the slope of $\mathrm{CAR}^{\mathrm{lvl}}_h$, the abnormal return from the previous close to rung $h$ in typical daily moves, on $V$ with day $\times$ tier fixed effects; $\kappa_h = 1$ is full convergence. Panel B: $\lambda_h$ is the share of the gap $U = V - G^{\mathrm{lvl}}$ closed by rung $h$, at D2 (from the close) and D1 (from 10:00); the half-life is the first rung with $\lambda_h \geq 0.5$ ($\infty$: not reached within $t{+}20$). Two-way clustered standard errors in parentheses; $t{+}k$ is the close of session $t+k$. ---: not estimated; $\lambda_h$ is reported at $t{+}1$, $t{+}5$ and $t{+}20$ from D2 and also at $C_t$ from D1, and the half-life belongs to $\lambda_h$ alone. $\kappa_h$ to two decimals, $\lambda_h$ to three, since it is an order smaller.
\item The anchor's readings are uninformative (near-constant outputs after one epoch); its rows are reported, not read as nulls. The API models fail the pre-registered contamination screens, so their results are model-specific.
\end{tablenotes}
\end{threeparttable}
\end{table}

\subsection{Calibration}\label{app:calibration}

\begin{figure}[htbp]
\centering
\includegraphics{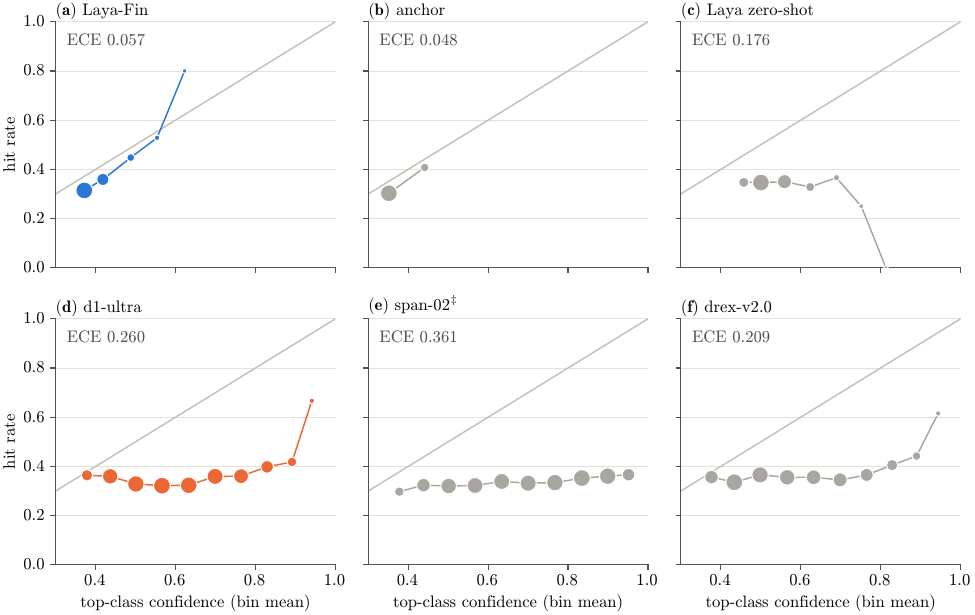}
\caption{\textbf{Laya-Fin's three-class forecasts stay close to the base rates (expected calibration error
\res{lf-ece}); the commercial models are overconfident (\res{drex-ece} to \res{span-ece}).} Reliability of the
pre-open reading's three classes (a fall, no move or a rise, split at $\pm 0.5$ typical moves) against the
five-day move $e_5$ on the test period: the hit rate of the top class by bin of its stated probability (15 bins;
the dot area grows with the number of events in the bin; the diagonal is perfect calibration). The anchor's
near-constant readings sit in two bins. $^{\ddagger}$See Appendix~\ref{app:prereg}.}\label{fig:a-reliability}
\end{figure}

Table~\ref{tab:a-calibration} gives the pre-registered calibration metrics of every model, and
Figure~\ref{fig:a-reliability} the reliability of the pre-open reading.

\begin{table}[htbp]
\centering
\small
\begin{threeparttable}
\caption{Calibration of the three-class forecasts: the trained models are calibrated but barely informative (Laya-Fin's ECE 0.057), the API models overconfident (ECE 0.209 to 0.361).}\label{tab:a-calibration}
\begin{tabular}{@{}>{\raggedright\arraybackslash}p{1.6in}S[table-format=1.3]S[table-format=-1.3]S[table-format=-1.3]S[table-format=1.2]S[table-format=1.2]S[table-format=1.3]S[table-format=1.3]@{}}
\toprule
Model, target & {ECE} & {Log-loss skill} & {Brier skill} & {AUC} & {$T$} & {ECE before} & {ECE after} \\
\midrule
Laya-Fin, $e_5$ & 0.057 & 0.001 & 0.000 & {---} & {---} & {---} & {---} \\
Laya-Fin, $z_{cc}$ & 0.033 & 0.010 & 0.010 & {---} & {---} & {---} & {---} \\
Laya-Fin, $|z_{cc}| \geq 1$ & {---} & {---} & {---} & 0.60 & {---} & {---} & {---} \\
Laya-Fin, $e_5$, temperature scaling & {---} & {---} & {---} & {---} & 1.00 & 0.058 & 0.057 \\
anchor, $e_5$ & 0.048 & 0.002 & 0.001 & {---} & {---} & {---} & {---} \\
anchor, $z_{cc}$ & 0.016 & 0.015 & 0.015 & {---} & {---} & {---} & {---} \\
anchor, $|z_{cc}| \geq 1$ & {---} & {---} & {---} & 0.58 & {---} & {---} & {---} \\
anchor, $e_5$, temperature scaling & {---} & {---} & {---} & {---} & 0.97 & 0.011 & 0.048 \\
Laya zero-shot, $e_5$ & 0.176 & -0.156 & -0.110 & {---} & {---} & {---} & {---} \\
Laya zero-shot, $z_{cc}$ & 0.179 & -0.177 & -0.120 & {---} & {---} & {---} & {---} \\
Laya zero-shot, $|z_{cc}| \geq 1$ & {---} & {---} & {---} & 0.52 & {---} & {---} & {---} \\
Laya zero-shot, $e_5$, temperature scaling & {---} & {---} & {---} & {---} & 6.33 & 0.176 & 0.110 \\
d1-ultra, $e_5$ & 0.260 & -0.194 & -0.194 & {---} & {---} & {---} & {---} \\
d1-ultra, $z_{cc}$ & 0.249 & -0.166 & -0.165 & {---} & {---} & {---} & {---} \\
d1-ultra, $|z_{cc}| \geq 1$ & {---} & {---} & {---} & 0.58 & {---} & {---} & {---} \\
d1-ultra, $e_5$, temperature scaling & {---} & {---} & {---} & {---} & 1.95 & 0.260 & 0.155 \\
span-02$^{\ddagger}$, $e_5$ & 0.361 & -0.460 & -0.361 & {---} & {---} & {---} & {---} \\
span-02$^{\ddagger}$, $z_{cc}$ & 0.342 & -0.410 & -0.315 & {---} & {---} & {---} & {---} \\
span-02$^{\ddagger}$, $|z_{cc}| \geq 1$ & {---} & {---} & {---} & 0.56 & {---} & {---} & {---} \\
drex-v2.0, $e_5$ & 0.209 & -0.135 & -0.138 & {---} & {---} & {---} & {---} \\
drex-v2.0, $z_{cc}$ & 0.204 & -0.119 & -0.120 & {---} & {---} & {---} & {---} \\
drex-v2.0, $|z_{cc}| \geq 1$ & {---} & {---} & {---} & 0.57 & {---} & {---} & {---} \\
\bottomrule
\end{tabular}
\begin{tablenotes}[flushleft]\footnotesize
\item ---: the source table has no value for the row; the statistic does not apply to it or was not computed.
\item The pre-open reading's three classes (a fall, no move or a rise, split at $\pm 0.5$ typical moves) against the targets; skill against val-A climatology; $T$ the temperature fit on val-A.
\item The anchor's readings are uninformative (near-constant outputs after one epoch); its rows are reported, not read as nulls.
\item $^{\ddagger}$span-02: after the freeze, a stability check that repeats earlier requests (the canary) fired for it; its build was unchanged, and production went ahead under a rule recorded in the deviations log.
\end{tablenotes}
\end{threeparttable}
\end{table}

\FloatBarrier
\section{Exploratory: the conditioning study}\label{app:exploratory}

\begin{figure}[htbp]
\centering
\includegraphics{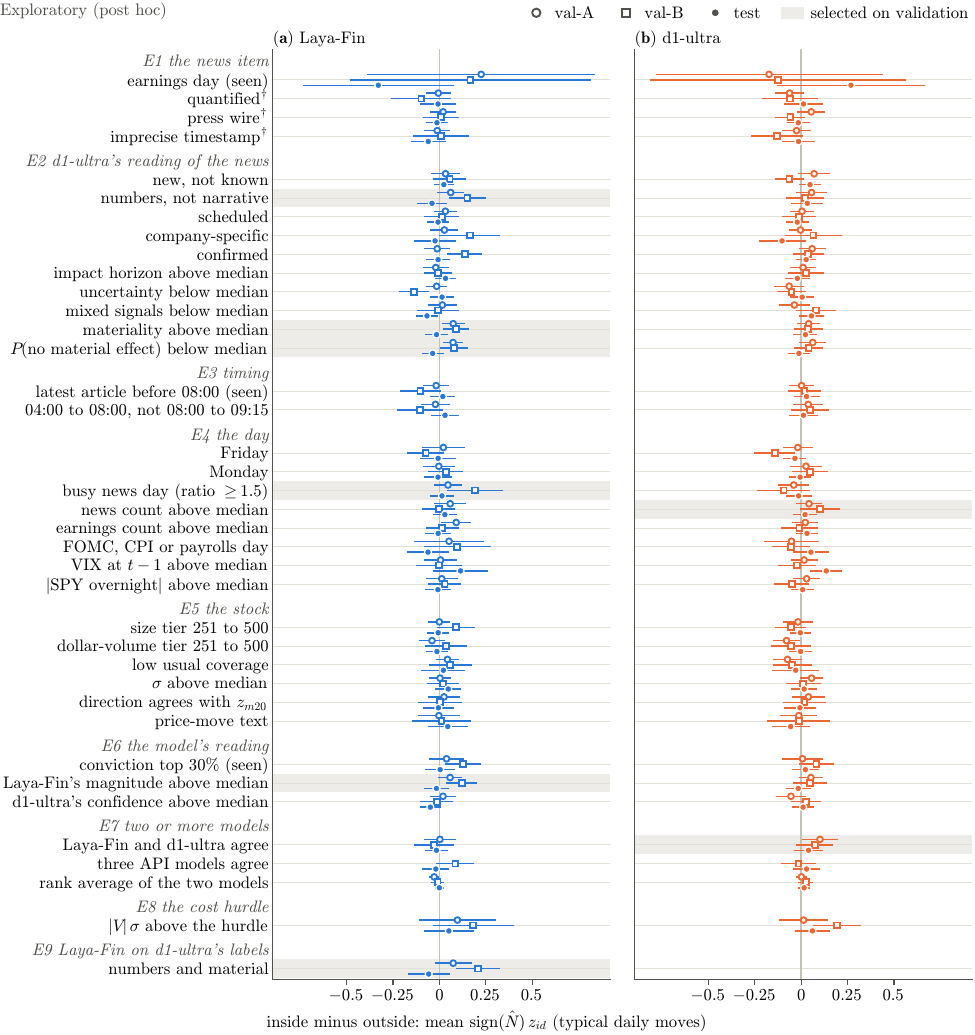}
\caption{\textbf{Exploratory (post hoc): none of the \res{cond-selected} conditions selected on two validation
quarters held on the test; \res{cond-reversed} changed sign.} For each (model, condition) pair, the difference
inside minus outside the condition of the mean $\mathrm{sign}(\Nhat)\cdot z_{id}$ (typical daily moves) on val-A
(circles), val-B (squares) and the test (dots), with 95\% intervals; shaded rows were selected by the rule fixed
before the test run (positive with $t \ge 1$ in both validation quarters and a pooled $t \ge 2$). $^{\dagger}$Vendor
fields are retrospective labels and only split events descriptively. ``(seen)'' marks conditions whose test values
the authors' assistant had seen before the study (Section~\ref{sec:exploratory}); none of them was
selected.}\label{fig:a-conditioning}
\end{figure}

Figure~\ref{fig:a-conditioning} shows every (model, condition) pair of the post hoc conditioning study of
Section~\ref{sec:exploratory}, with the \res{cond-selected} pairs that the validation rule selected shaded. It is
exploratory:
the study was requested after unblinding, and the figure supports no claim.

\section{The replication package}\label{app:replication}

The project's repository, \url{https://github.com/AmirDehkordi/NewsAction}, is the replication package. It holds
the pre-registration with its amendments and its full deviations log, the freeze record, the code of every stage
with its tests, the committed outputs and a report of every stage, the ledgers of trials, spend and test-data
access, and the sources of this paper, whose tables, figures and numbers one command rebuilds from the committed
outputs without raw data, an API or a GPU (two strategy tables, which need position-level files kept with the
data, are carried over as committed). The data are private and are not part of the package: the news text is
licensed and is not redistributed, and neither are the price data or the index membership.

\end{document}